\documentclass{article}

\usepackage{arxiv}
\usepackage[utf8]{inputenc}
\usepackage{graphicx}
\usepackage{amsmath}
\usepackage{amssymb}
\usepackage{array}
\usepackage{enumerate}
\usepackage{enumitem}
\usepackage{algorithm}
\usepackage{algpseudocode}
\usepackage{caption}
\usepackage{subcaption}
\usepackage{booktabs}
\usepackage{multirow}
\usepackage{svg}
\usepackage{times}
\usepackage{soul}
\usepackage{url}
\usepackage{float}
\usepackage{flafter}
\usepackage{latexsym}
\usepackage{color}

\DeclareMathOperator*{\argmax}{arg\,max}
\DeclareMathOperator*{\argmin}{arg\,min}
\newcommand {\otoprule}{\midrule [\heavyrulewidth]}
\newcolumntype {+}{ >{\global\let\currentrowstyle\relax}}
\newcolumntype {^}{ >{\currentrowstyle }}
 \newcommand {\rowstyle}[1]{\gdef\currentrowstyle{#1} %
 #1\ignorespaces
 }
\newcommand{\tabhead}{\rowstyle{\bfseries}}

\title{Explainable Bayesian Optimization}

\author{
 Tanmay Chakraborty \\
  Continental Automotive Technologies GmbH, AI Lab Berlin, Germany\\
  University of Marburg
  Marburg, Germany\\
  \texttt{tanmay.chakraborty@continental-corporation.com} \\
   \And
 Christian Wirth \\
  Continental Automotive Technologies GmbH, Frankfurt, Germany \\
  \texttt{christian.2.wirth@continental-corporation.com} \\
  \And
 Christin Seifert \\
  University of Marburg
  Marburg, Germany \\
  \texttt{christin.seifert@uni-marburg.de} \\
}

\begin{document}
\maketitle

\begin{abstract}
Manual parameter tuning of cyber-physical systems is a common practice, but it is labor-intensive. Bayesian Optimization (BO) offers an automated alternative, yet its black-box nature reduces trust and limits human-BO collaborative system tuning.  Experts struggle to interpret BO recommendations due to the lack of explanations. This paper addresses the post-hoc BO explainability problem for cyber-physical systems. We introduce TNTRules (Tune-No-Tune Rules), a novel algorithm that provides both global and local explanations for BO recommendations. TNTRules generates actionable rules and visual graphs, identifying optimal solution bounds and ranges, as well as potential alternative solutions.  Unlike existing explainable AI (XAI) methods, TNTRules is tailored specifically for BO, by encoding uncertainty via a variance pruning technique and hierarchical agglomerative clustering. A multi-objective optimization approach allows maximizing explanation quality. We evaluate TNTRules using established XAI metrics (Correctness, Completeness, and Compactness) and compare it against adapted baseline methods. The results demonstrate that TNTRules generates high-fidelity, compact, and complete explanations, significantly outperforming three baselines on 5 multi-objective testing functions and 2 hyperparameter tuning problems.  
\end{abstract}


\section{Introduction}\label{sec:introduction}

Manual parameter tuning of cyber-physical systems by experts has been the industry norm~\cite{neumann2019data,nagataki2022online}. To reduce the manual effort, Bayesian Optimization (BO) has been introduced for parameter tuning~\cite{shahriari2015taking}. BO is a model-based optimization technique with a Gaussian Process (GP) backbone that returns black-box recommendations. Since the recommendations provided by BO may not be accurate, due to simplified objectives and approximation errors, the parameter tuning process cannot be fully automated and still requires expert intervention~\cite{sundin2022human,10.1007/978-3-030-43651-3_58}. 

Typically, experts in domains such as automotive tuning, engine calibration, laser alignment, and other hardware-oriented fields have limited computer science backgrounds. Therefore, they primarily use BO results in a post-hoc manner (experts not involved in the BO development process), usually as a tool for fine-tuning physical systems~\cite{10.1007/978-3-662-64408-9_7,neumann2019data}. In addition, complex industrial design processes for aerospace components~\cite{liu2022airfoil}, mechanical bearing designs for fault tolerance~\cite{ortiz2024enhanced}, as well as instrumentation designs in particle accelerators can also use BO as a tool to explore design spaces, making the design more autonomous and easier~\cite{roussel2024bayesian}. In this case, collaboration refers to receiving recommendations from BO and using them on the hardware system.

However, BO is a black box algorithm. Thus, it only returns suggested parameter settings without explanations, which hinders the expert's trust in the tuning process~\cite{rodemann2024explaining,adachi2024looping,chakraborty_post-hoc_nodate}. This interpretability challenge in BO mirrors similar challenges in AI that led to the rise of Explainable AI (XAI)~\cite{rudin2022black}. While several techniques have been developed to explain AI decisions, explanations for optimization algorithms are rare. 
Explainable Bayesian Optimization aims to make BO recommendations transparent by providing explanations to experts. Explanations for BO would instill trust in the BO recommendations and provide a better collaborative experience for the expert~\cite{rodemann2024explaining,adachi2024looping,chakraborty_post-hoc_nodate}. The main goal is to elaborate the design spaces of the optimization problem, as learned by BO, using simple and easy to understand rules, so that users can use the recommendations in an interpretable way.

\begin{figure}[t]
    \centering
    \includegraphics[width=\columnwidth]{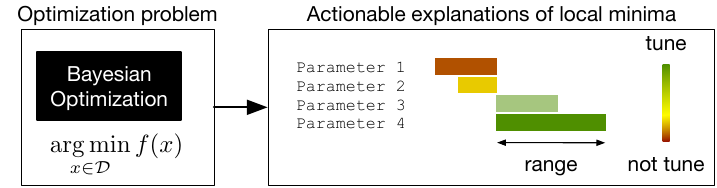}
    \caption{Explainable Bayesian Optimization with actionable local explanation for one minimum (multiple minima would have multiple graphs): red/yellow bars indicate optimal parameters, green bars indicate tunable parameters. Bar lengths indicate tuning ranges.}
    \label{fig:xboteaser}
\end{figure}

In this paper, we address the post-hoc BO explainability problem. More specifically, we present a method for explaining BO recommendations to experts who calibrate and tune cyber-physical systems (Fig.~\ref{fig:xboteaser}). Based on the literature, we identified two main requirements for providing these explanations~\cite{chakraborty_post-hoc_nodate}: i) identifying optimal solution bounds and ranges for tuning, and ii) identifying and providing bounds and ranges for tuning potential alternative solutions (in the case of local minima).

Current XAI methods are designed to explain machine learning models, such as classification and clustering algorithms, and do not meet the requirements~\cite{schwalbe2023comprehensive}. We address this gap by introducing TNTRules (Tune-No-Tune Rules), a post-hoc rule-based explanation algorithm for BO. TNTRules provides local explanations through visual graphs and actionable rules that indicate which parameters should be adjusted or left unchanged to improve results. For global explanations, TNTRules generates a ranked ``IF-THEN'' rule list. We use Hierarchical Agglomerative Clustering (HAC) to identify rules within the optimization space, including their associated uncertainties~\cite{ran2023comprehensive}.
Unlike many XAI methods, TNTRules is controllable, allowing adjustments for high-quality explanations. Controllability means adjusting the number of rules mined. In BO, this means changing the size of the bounding boxes that divide the optimization space, which affects explanation quality. In terms of explanation, it means allowing the user to have different views on the optimization space. We provide high-quality explanations through multi-objective optimization, aiming to maximize explanation quality by automatically tuning a threshold $t_s$ and balancing different quality metrics.

We evaluate the explanation quality (qualitative and quantitative) of TNTRules using metrics derived from the XAI literature (Co-12 framework)~\cite{10.1145/3583558}. Although there are numerous, competing metrics for evaluating explanation quality, we focus on the three evaluation criteria from the Co-12 framework, that are relevant to our use case: Correctness (quantitative), Completeness (quantitative and qualitative), and Compactness (qualitative). Ideally, explanations should be consistent with the model they are explaining (correctness), thorough in their coverage (completeness), and concise enough for users to easily understand (compactness). A high-quality explanation would score well on all three metrics.

Current literature lacks directly comparable methods for evaluating TNTRules -- apart from RXBO~\cite{chakraborty_post-hoc_nodate} -- which is why we adapted existing approaches (decision rules~\cite{apte1997data} and RuleXAI~\cite{macha2022rulexai}) from the XAI literature to provide diverse baselines for comparison. Our results show that TNTRules generate high-fidelity, compact, and complete explanations, which shows substantial improvement from the baseline methods.
More specifically, we make the following contributions:
\begin{enumerate}
\item We introduce TNTRules, a method for explaining Bayesian Optimization as a set of global rules and local actionable explanations.
\item We introduce a variance pruning technique for hierarchical agglomerative clustering to encode uncertainty.
\item We introduce an optimizaton method based on multi-objective criteria to maximize the quality of explanations.
\item Our functionally grounded evaluation of hyperparameter optimization tasks shows that TNTRules outperform other explanation methods. On common optimization test functions, we show TNTRules' capability to identify the optimal region and its potential to detect other local minima.
\end{enumerate}

\section{Related Work}\label{sec:relatedwork}
This section provides an overview of XAI, explanation presentations, and evaluation of the XAI method. This is followed by rule-based explanations and optimizing the quality of explanations.

\subsection{Explainable AI}

XAI can be broadly classified into transparent models and post hoc explanation methods, which are either specifically designed for certain model classes or applicable to a wide range of models (model agnostic)~\cite{guidotti2018survey}. 
The resulting explanations can be global, explaining the entire model, or local, explaining a single sample. TNTRules generates a global surrogate model~\cite{craven1995extracting} as a set of rules and local actionable rules. 

\textbf{Explanations for Bayesian Optimization} are either post-hoc, i.e. explaining the minima and their bounds \cite{chakraborty_post-hoc_nodate}, or online, i.e. explaining BO's next point selection strategy~\cite{adachi2024looping,rodemann2024explaining}. There has been additional work on explaining hyperparameter optimization tasks~\cite{moosbauer2024,segel23a}. These works are not directly comparable to ours because they i) do not explain the black-box function in BO, ii) do not address or bound minima in the optimization space, iii) do not provide tuning ranges, and iv) are not rule-based systems.

The \textbf{post-hoc method} of~\cite{seitz2022gradient} explains the core of BO, i.e., Gaussian Process (GP), primarily by producing feature rankings, possibly uncertainty-aware~\cite{seitz2022gradient}, but they do not meet the explainable Bayesian optimization's requirements i)-ii). RXBO, the closest work to ours, introduces a rule miner for optimization, but does not deal with uncertainty during clustering and lacks multiple minima identification~\cite{chakraborty_post-hoc_nodate}. We take inspiration from the literature on rule-based surrogate models~\cite{coppens2019distilling,murdoch2017automatic,i2022consistent}. These methods distill a base model into sets of rules by using a (probabilistic) surrogate model. 
\textbf{Online methods} are suitable for domains such as drug discovery, materials exploration, and manufacturing, where experts rely heavily on computer simulations for discovery processes~\cite{frazier2016bayesian,colliandre2023bayesian}. In such domains, understanding BO's next point selection (while the algorithm is running) helps experts make informed decisions. Unlike online methods, post-hoc methods (this work and~\cite{chakraborty_post-hoc_nodate}) focus on the user group that uses BO results in a post-hoc manner (not involved in the algorithm running phase, but using BO as a tool)~\cite{10.1007/978-3-662-64408-9_7}. 

Explanation generation is typically tightly coupled to a \textbf{presentation format}. Following the model inspection pipeline of \cite{guidotti2018survey}, we argue that explanation generation and visualizations are distinct aspects, emphasizing the central role of explanation presentation in improving understandability. TNTRules translates rules into visual explanations, following design patterns from popular visual-centric approaches such as SHAP~\cite{lundberg2019explainable} and LIME~\cite{ribeiro2016why}.
Similar to evaluating machine learning models, explanation methods need to be evaluated. \textbf{XAI evaluation} libraries such as Quantus~\cite{hedstrom2023quantus} are actively involved in establishing evaluation benchmarks. A comprehensive study~\cite{10.1145/3583558} has identified 12 properties for evaluating XAI. We adopt an appropriate subset of these properties to evaluate our TNTRules. We use functionally based evaluations, which is standard in XAI~\cite{vilone2021quantitative,zhou2021evaluating}.   

\subsection{Rule Mining and Rule-based Explanations}
Association rule mining has many established algorithms such as Apriori, and Eclat~\cite{yazgana2016literature}. XAI has adopted them for rule-based explanations because of their transparency, fidelity, and ease of use~\cite{van2021evaluating}. 
We take a similar stance, choosing rules for their high level of comprehensibility. Similar to~\cite{wang2015falling}, we present rules in descending order of utility. TNTRules are not fuzzy rules, meaning rules do not have a degree of uncertainty in themselves, but they represent the uncertainty of the GP model. The rules maintain precision within the domains they define, ensuring clarity and accuracy in the explanations provided~\cite{Magdalena2015}.

\subsection{Optimizing Quality of Explanations}
Controllability, i.e., mechanisms for end users to control the explanation method, is a desideratum for explanations~\cite{10.1145/3583558}. Controlling an XAI method means being able to define and set parameters for an explanation method~\cite{fernandes2022learning}. Many XAI methods, such as SHAP and LIME, lack control and assume that the explanations they produce are consistently optimal, making them black boxes explaining black boxes (the machine learning models).
The existing literature on controllable XAI methods is limited. A recent paper~\cite{pahde2023optimizing} introduced the concept of hyperparameter search for XAI. Building on this foundation, our contribution advances the field by proposing a novel approach to XAI hyperparameter search using multi-objective optimization.

\begin{figure*}[htb]
    \centering
    \includegraphics[width=\columnwidth]{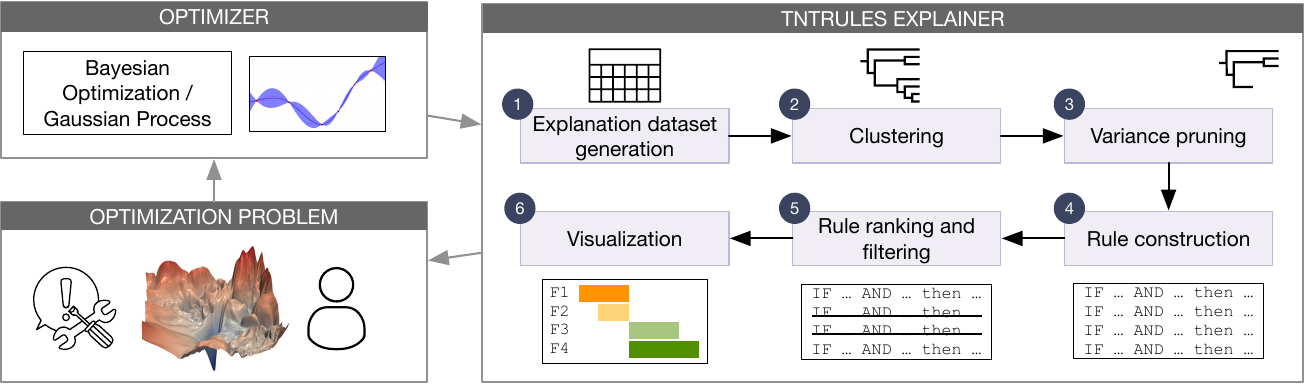}
    \caption{Given an optimization problem, TNTRules explains the optimizer by identifying parameter bounds for optimal regions in the search space. TNTRules first generates an explanation dataset (1) by sampling the search space. The dataset is clustered (2); these clusters are pruned (3), and rules are constructed for them (4). These rules are then ranked and filtered (5), and the final set is presented as explanations to the end user (6).\protect\footnotemark}
    \label{fig:xboarch}
\end{figure*}
\footnotetext{3D loss landscape figure credited to: \url{https://github.com/tomgoldstein/loss-landscape}}

\section{Background and Problem Setting}\label{sec:background}
This section briefly describes the background of Bayesian Optimization (BO) and Gaussian Processes (GPs) relevant to this paper, followed by the problem setup.

\textbf{Bayesian Optimization}, a sequential model-based optimization technique, uses a probabilistic model, usually a Gaussian Process (GP). BO uses uncertainty-aware exploration/exploitation trade-offs to reduce the required number of iterations~\cite{garnett_bayesoptbook_2023}.

BO identifies minima ($\mathbf{x_{opt}}$) for a black-box objective $f(\mathbf{x})$ defined in a bounded search-space $\mathcal{D}\subseteq\mathcal{R}^d$:
\begin{align}\label{eq:1}
\mathbf{x_{opt}} = \argmin_{\mathbf{x} \in \mathcal{D}} f(\mathbf{x}). 
\end{align}
$f$ is a black-box function that is \textit{expensive to evaluate, noisy, and not known in closed form.} The GP serves as an approximator of the objective function~\cite{shahriari2015taking}. We assume that the underlying GP approximates the optimization space sufficiently. Minimizing the GP approximation error is a separate research area~\cite{rodemann2022accounting}.

A \textbf{Gaussian Process} is a set of random variables with each finite set of those variables following a multivariate normal distribution. The distribution of a GP is the joint distribution of all variables. In a GP, a distribution model is formulated around functions, each having a mean $m(\mathbf{x})$ and covariance function, commonly referred to as the kernel function $k(\mathbf{x},\mathbf{x'})$. These collectively dictate the behavior of each function $f(\mathbf{x})$ at the specific location $\mathbf{x}$. When the mean is defined as $m(\mathbf{x}) = \mathbb{E}[f(\mathbf{x})]$, and the kernel function is articulated as $k(\mathbf{x},\mathbf{x'}) = \mathbb{E} [(f(\mathbf{x}) - m(\mathbf{x}))(f(\mathbf{x'}) - m(\mathbf{x'})]$, the GP framework is formally denoted as $f(\mathbf{x}) \sim GP (m(\mathbf{x}), k(\mathbf{x},\mathbf{x'}))$.~\cite{Rasmussen2005-yj}

\textbf{Hierarchical Agglomerative Clustering (HAC)}~\cite{murtagh2012algorithms} groups data points into clusters by iteratively merging the closest pairs, starting as individual clusters and visualized via a dendrogram. The linkage matrix records each merger, noting the merged clusters, their distance, and the new cluster size. Ward's criterion, a variant of HAC, minimizes within-cluster variance by merging clusters that least increase the total variance, producing compact clusters. This is quantified as $\Delta V(A, B) = \frac{|A| \cdot |B|}{|A| + |B|} \cdot \| \bar{A} - \bar{B} \|^2$, where $|A|$ and $|B|$ are the sizes, and $\bar{A}$ and $\bar{B}$ the centroids of clusters A and B~\cite{ward1963hierarchical,murtagh2011ward}.

\textbf{Problem setting}: Given the objective function in Equation~\eqref{eq:1}, the boundaries of the search space $\mathcal{D} = \{\mathbf{x} \in \mathcal{R}^d: lb^j \leq x^j \leq ub^j, \forall j \in {1,.., d}\}$, and the GP model, the post-hoc explanation goal is to find rules that bound the minima of this function. An explanation is defined as rules bounding minima of the objective function. These rules must be constrained within the search space, considering the variance of inputs and outputs as modeled by the GP. Let's denote input points that are neighboring minima by a variance $v$ as $\mathbf{X_v}$,  the GP mean and standard deviation as $\boldsymbol{\mu}$ \& $\boldsymbol{\sigma}$ respectively. An explanation for minima takes the form: $[min(X_v), max(X_v)] => [min(\mu - 2 * \sigma), max(\mu + 2 * \sigma)]$. This indicates that the bounds of the rule are determined by the minimum and maximum input points, which are grouped based on their variance, with a result representing the confidence interval for the corresponding inputs. Input points that exceed the variance threshold should be separate rules. These rules would need to be checked for relevance, i.e. whether they contain additional local minima. Rules that do not contain a likely minimum must be filtered out. The bounds of the rules provide the expert with tunable ranges for each parameter that can be tuned to maintain a certain level of utility.  In case of multiple, local minima similar ranges should be provided for each. 

\begin{figure*}[htb]
    \centering
    \includegraphics[width=\columnwidth]{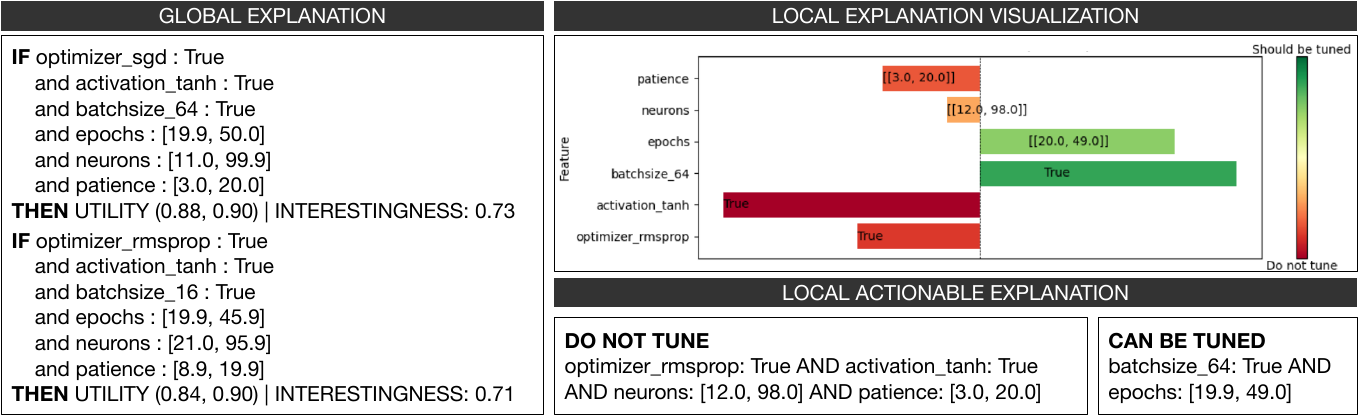}
    \caption{Example of the three forms of explanations generated using the HPO use case with the MLP model. Global explanations consist of 22 rules; here, we present two as examples. Local explanations include a sensitivity graph and actionable insights detailing which parameters to tune and what ranges.}
    \label{fig:ruleview}
\end{figure*}

\section{TNTRules (Tune-No-Tune Rules)}\label{sec:tntrules}
TNTRules (Fig.~\ref{fig:xboarch}) is our algorithm (Algorithm~\ref{tntrules}) for extracting and visualizing BO solution spaces. In TNTRules, we create an explanation dataset (1) for the objective function $f$, cluster it (2), and prune the cluster tree (3), then generate rules describing the clusters (4), rank and filter them for final explanations (5), and visualize the rules in a user-friendly manner (6) (cf. Fig.~\ref{fig:ruleview}). Additionally,  TNTRules is optimized with multi-objective optimization to produce high-quality explanations (cf. Sec~\ref{sec:optimization}).

\begin{algorithm}[htb]

\caption{TNTRules Algorithm}\label{tntrules}
\scriptsize
\begin{algorithmic}[1]
\Require {BO search space:~$\mathcal{D},$ \\
No. of explanation samples:~$N_e,$ \\
Clustering threshold:~$t_s$} \\
Interestingness threshold:~$t_\alpha$\\
BO backbone GP: $GP$
\Procedure{$TNTRules$}{$\mathcal{D} ,N_{e}, t_s$}
\State \textbf{\# Explanation dataset generation (Sec.~\ref{sssec:expdataset})}
\State $\mathbf{X_{e}} \gets \{x_1, \ldots, x_{N_e}\} \sim \mathcal{U}(\mathcal{D})$ 
\State $\boldsymbol{\mu}, \boldsymbol{\sigma_y} \gets GP_{predict}(\mathbf{X_{e}})$
\State $\mathbf{E} \gets [\mathbf{X_{e}} ;\boldsymbol{\mu} ; \boldsymbol{\sigma_y}]$ 
\State \textbf{\#Clustering (Sec.~\ref{sssec:clustering}), Variance pruning (Sec.~\ref{sec:pruning}), and Rule construction (Sec.~\ref{sssec:ruleconst})}
\State $\mathbf{E_{link}} \gets Clustering(\mathbf{E})$
\State $\mathbf{K} \gets VariancePruning(\mathbf{E_{link}}, t_s)$
\State $\rho_{\dashv}, \rho_{\vdash} \gets RuleConstruction(\mathbf{E}, \mathbf{K})$ 
\State \textbf{\# Rule ranking and filtering (Sec.~\ref{sssec:rulerank})}
\For {$i \  in\  \rho_{\dashv}$}
\State $\rho_{temp}^i \gets find(\mathbf{X_{e}} \in [\rho_{\dashv}^i ; \rho_{\vdash}^i])$ 
\State $Rel^i \gets max(likelihood(GP_{predict}(\rho_{temp}^i)))$ 
\State $Covr^i \gets ECDF(\rho_{\dashv}^i, \mathbf{X_{e}})$
\State $Supp^i \gets ECDF([\rho_{\dashv}^i ; \rho_{\vdash}^i], [\mathbf{X_{e}}; \boldsymbol{\mu}])$ 
\State $Con^i \gets Supp^i/Covr^i$ 
\State $\alpha^i \gets weightedSum(Rel^i,Covr^i, Supp^i, Con^i)$ 
\EndFor
\State $\rho \gets FilterRules([\rho_{\dashv}^i ; \rho_{\vdash}^i], \alpha, t_\alpha)$ 
\State \textbf{return} $\rho$
\EndProcedure
\end{algorithmic}
\end{algorithm}

\subsection{Explanation Dataset Generation}\label{sssec:expdataset}

In standard XAI settings, the explanation method has access to a data set. Usually, this is the training data, the ground-truth labels, and the model's predictions. In the BO setting, the data set does not exist a-priori. BO dynamically acquires data during execution from the search space $\mathcal{D}$~\cite{shahriari2015taking}. The sampled data is small in quantity and biased toward optimal areas (BO iteratively refines the objective function while avoiding exhaustive sampling of the optimization space). Thus, the sampled data is unsuitable as an explanation dataset.

Therefore, we generate an explanation dataset by uniformly sampling the search space for ($N_e$) samples $\mathbf{X_{e}} = \{x_1,\ldots, x_{N_e}\} \sim \mathcal{U}(\mathcal{D})$ (algorithm lines 6-7). Our explanation dataset then has a comprehensive view of the optimization space. Sampling the search space is less computationally expensive than acquiring data during execution. In this setting, we assume that the GP sufficiently approximates the black-box function during the optimization run (cf. Sec~\ref{ssec:ablation}). The GP is queried with $\mathbf{X_{e}}$ to infer the posterior distribution i.e., mean $\boldsymbol{\mu}$ and standard deviation $\boldsymbol{\sigma_y}$. The resulting explanation dataset is formulated as $\mathbf{E} = [\mathbf{X_{e}} ; \boldsymbol{\mu} ; \boldsymbol{\sigma_y}]$ (algorithm line 8).

\subsection{Clustering}\label{sssec:clustering}

We use clustering to identify significant regions in the approximated posterior distribution ($\mathbf{E}$), considering the uncertainty within each region. Using the Ward criterion, we apply HAC to the explanation dataset $\mathbf{E}$ (algorithm line 10). The resulting linkage matrix $\mathbf{E_{link}}$ consists of links and distances between clusters. $\mathbf{E_{link}}$ is of size ($N_e - 1$), where each row corresponds to a merge or linkage step in the clustering process. The resulting clustering can be visualized in the form of a dendrogram which represents the hierarchical tree structure of the linkage matrix. Pruning at different distances/levels of $\mathbf{E_{link}}$ produces clusters with varying coverage~\cite{salvador2004determining}. It is a challenge to determine the optimal clustering threshold ($t_s$) or pruning distance for pruning that produces meaningful explanations, i.e., clusters that are both localized and cover areas around minima (cf. Sec~\ref{ssec:tuningcutoff}).

\subsection{Variance Pruning}
\label{sec:pruning}
Instead of using a traditional dendrogram threshold to merge clusters, we employ the HAC method to obtain only the linkage structure $\mathbf{E_{link}}$, followed by a distance-based pruning mechanism. However, conventional distance metrics often fail to separate clusters effectively in cases of high variability or outliers~\cite{KLUTCHNIKOFF2022105075}.

To overcome this, we propose a variance-based pruning approach that captures data uncertainty through the linkage matrix (cf. Sec~\ref{ssec:ablation}). Variance reflects uncertainty in BO, making it a meaningful metric for clustering. The intuition here is that highly uncertain areas would require more human attention and thus must be separated from very certain areas.

Let $\boldsymbol{\mu}$ represent the target values of the leaf nodes (produced by the GP), $n$ the number of leaf nodes, and $t_s$ a predefined clustering threshold. The merging condition is:

\[
\begin{cases}
\text{Merge leaf nodes} & \text{if } \text{Var}(\boldsymbol{\mu}) \leq t_s \\
\text{Do not merge leaf nodes} & \text{if } \text{Var}(\boldsymbol{\mu}) > t_s
\end{cases}
\]

Here, $\text{Var}(\boldsymbol{\mu})$ denotes the variance of $\boldsymbol{\mu}$, and $t_s$ serves as a hyperparameter controlling cluster formation.

\subsection{Rule Construction}\label{sssec:ruleconst}
Based on the formed clusters $\mathbf{K}$, we extract the data points $\mathbf{X^i_e}$ belonging to the subtree/cluster from $\mathbf{X_e}$ to generate rules for each cluster. The antecedent's rule bounds ($\rho_{\dashv}$) are determined by taking the minimum and maximum over the extracted data points represented by each subtree. Similarly, for the consequent part ($\rho_{\vdash}$), we calculate the confidence interval for the corresponding GP posterior, capturing the uncertainty: $\boldsymbol{\mu^i} \pm 2 * \boldsymbol{\sigma^i_y}$ (algorithm line 12). A single rule looks like $[min(X^i_e), max(X^i_e)] => [min(\mu^i - 2 * \sigma^i_y), max(\mu^i + 2 * \sigma^i_y)]$. This step produces the full rule set denoted as $\boldsymbol{\rho_{all}}$. 

\subsection{Rule Ranking and Filtering}\label{sssec:rulerank}
The initial rule set $\boldsymbol{\rho_{all}}$ must be filtered and ranked based on their quality for producing the final explanations. We employ a quantitative framework to assess rule quality, focusing on localization within the optimization space and the ability to cover meaningful regions around potential solutions. Our evaluation integrates four metrics: 
\begin{enumerate}
    \item Coverage ($Covr$): how much a rule's antecedent covers a region within the search space relative to the total area (algorithm line 17).
    \item Support ($Supp$): evaluates the alignment of a rule's domain with regions containing actual data (algorithm line 18).
    \item Confidence ($Con$): is the ratio of Support to Coverage (algorithm line 19).
    \item Relevance ($Rel$): is the maximum log-likelihood derived from the GP applied to the subset of data that conforms to a particular rule. Let $\mathbf{l_s}$ denote the log-likelihood values for such instances, and $\mathbf{l}$ represent the log-likelihoods of the entire dataset $Rel = max_{\mathbf{l_s} \in \mathbf{l}}(\mathbf{l_s})$ (algorithm line 16) ~\cite[Chapter 2]{Rasmussen2005-yj}.
\end{enumerate}

Coverage, and Support are standard in rule mining and computed using an empirical cumulative probability distribution (ECDF)~\cite{WITTEN201767}. While effective in measuring the quality of the generated bounding boxes, they have limitations in locating optimal solutions. We introduce the Relevance metric to address this limitation. Relevance effectively identifies regions where the GP is most likely to locate optimal solutions.

Finally, we define the overall interestingness of a rule, bounded between (0,1], as the weighted sum: $\alpha = w_1*Covr + w_2*Supp + w_3*Con + w_4*Rel$, higher is better (algorithm line 20). The most interesting rules are presented by filtering the rule set $\boldsymbol{\rho_{all}}$ based on the interestingness threshold $t_\alpha$ resulting in explanations $\boldsymbol{\rho}$ (algorithm line 22).

\subsection{Visualization}\label{sssec:vizual}

TNTRules presents explanations $\boldsymbol{\rho}$ to the recipient through three visualization modes: global textual rules, local visual graphs, and local actionable explanations. The global explanations are expressed as an ordered list of ``IF-THEN" rules, organized in decreasing order of interestingness $\alpha$ (cf. Fig.~\ref{fig:ruleview}, left). Textual rules provide a simpler and interpretable model but are challenging to comprehend for an explanation recipient, especially in scenarios with long antecedents. Local explanations are given as a rule extracted from our global rule set that explains a particular sample. This allows for easier comprehension than a full rule set.

In Fig.~\ref{fig:ruleview}, local explanation visualizations appear on the right. To generate these graphs, we perform a sensitivity analysis of the GP model~\cite{xu2008uncertainty}, focusing on data samples covered by a rule. The sensitivity analysis identifies optimization model parameters that, when adjusted, can improve results. Tailored to the use case, we generate local, actionable explanations that cover all the input parameters of the problem. This representation abstracts non-essential rule components during tuning, improving user understanding. The graph and actionable explanations guide the user by indicating parameters that require attention and those that can be ignored, simplifying interpretation, especially in cases with long antecedents. Aptly, these rules are called ``TUNE-NO-TUNE'' rules. The interpretation of the local actionable explanation given in the figure is as follows: do not tune the parameters $optimizer\_rmsprop$, $activation\_tanh$, $neurons$, and $patience$ because they are already in their optimal state according to the backbone GP model. While tuning is required from your side for parameters $batchsize\_64$ and $epoch$, this means that for this MLP model, a good batch size is not 64 but some other value from the search space, and also the epoch value can be adjusted from 19 to 49 for better results.

\section{Illustrative Example}

\begin{figure*}[!tb]
     \centering
     \begin{subfigure}[b]{0.49\textwidth}
         \centering
          \includegraphics[width=\linewidth]{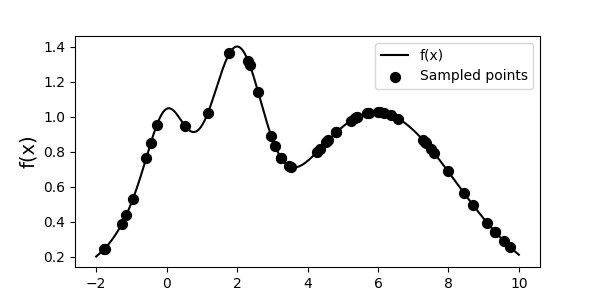}
         \caption{Sampling of the explanation dataset $\mathbf{X_{e}}$}
     \end{subfigure}
     \begin{subfigure}[b]{0.49\textwidth}
         \centering
         \includegraphics[width=\linewidth]{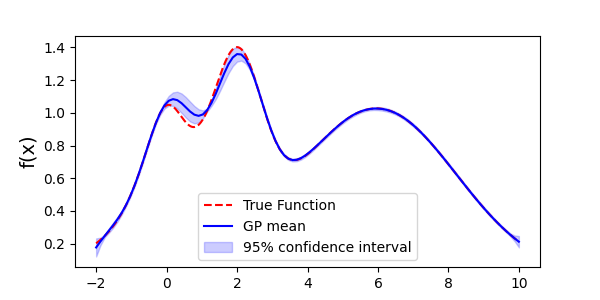}
         \caption{Prediction of black-box function $GP(\mathbf{X_{e}})$}
     \end{subfigure}
     \begin{subfigure}[b]{\textwidth}
         \centering
         \includegraphics[width=\linewidth]{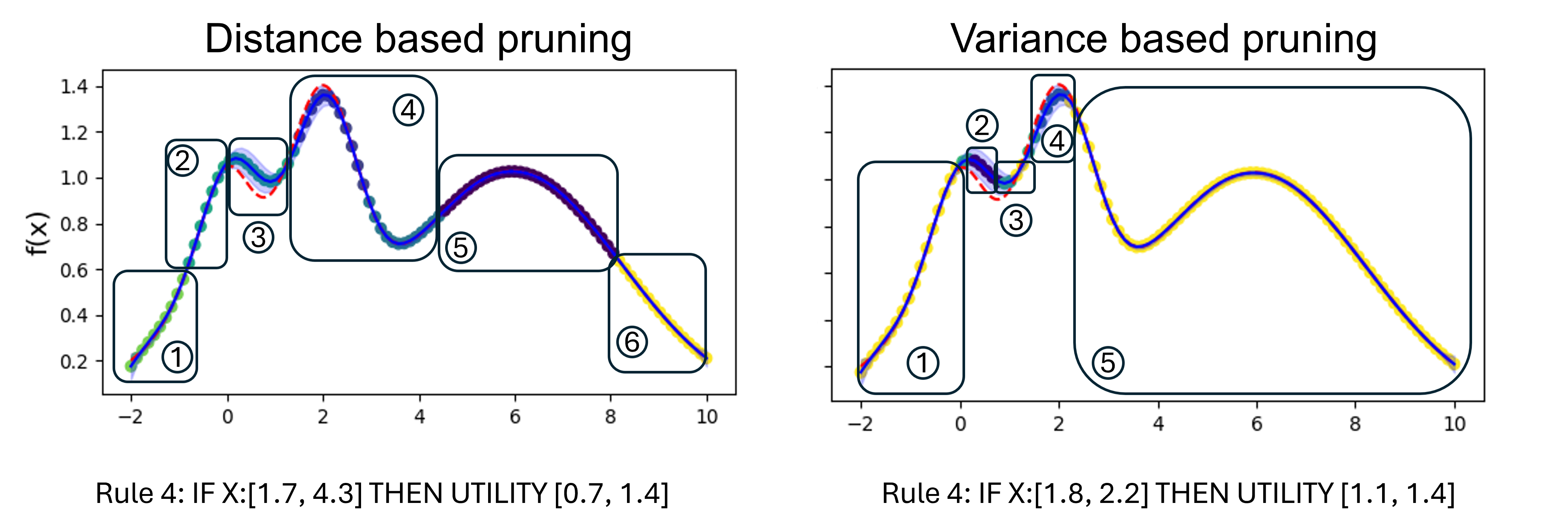}
         \caption{HAC based clustering and distance and variance pruning of explanation dataset $\mathbf{E}$, along with rule explanation for the optimal area (cluster 4)}
     \end{subfigure}

  \caption{(a) Generate X samples for the explanation dataset $\mathbf{X_{e}}=40$ from the search space $\mathcal{D}$. (b) Apply the GP model, trained to approximate our optimization target $f(x) = e^{-(x-2)^2} + e^{-(x-6)^2/10} + \frac{1}{x^2 + 1}$, to obtain predictions $\boldsymbol{\mu}$ and $\boldsymbol{\sigma_y}$. (c) Form the explanation dataset $\mathbf{E} = [\mathbf{X_{e}}, \boldsymbol{\mu}, \boldsymbol{\sigma_y}]$ and cluster it with HAC. To have meaningful clusters, prune the hierarchical structure. Finally, generate rules as an explanation for the different clustered areas.}
    \label{fig:samplingfunction}
\end{figure*}

We provide an illustrative example for TNTRules in Fig.~\ref{fig:samplingfunction}. 
The first step is to uniformly sample $\mathbf{X_{e}}=40$ points from $\mathcal{D}$, visualized as the x-axis in Fig.~\ref{fig:samplingfunction}a. We then apply the GP model, trained to approximate our optimization target $f(x) = e^{-(x-2)^2} + e^{-(x-6)^2/10} + \frac{1}{x^2 + 1}$, to obtain predictions $\boldsymbol{\mu} ; \boldsymbol{\sigma_y}$. This is visualized in Fig.~\ref{fig:samplingfunction}b. (cf. Alg.~\ref{tntrules} and Sec.~\ref{sssec:expdataset}). 

Subsequently, the explanation dataset $\mathbf{E} = [\mathbf{X_{e}} ; \boldsymbol{\mu} ; \boldsymbol{\sigma_y}]$ is formed and clustered using HAC (Sec.~\ref{sssec:clustering}). Following, distance- and variance-pruning are employed (Sec.~\ref{sec:pruning}), forming different clusters as given by different bounding boxes in Fig.~\ref{fig:samplingfunction}c.

Finally, we construct rules from the bounding boxes (Sec.~\ref{sssec:ruleconst}), which are then ranked based on how interesting they might be to a user (Sec.~\ref{sssec:rulerank}), rule-based explanations or visual explanations are generated for the most interesting rules, and the rest are filtered out (Sec.~\ref{sssec:vizual}) Fig.~\ref{fig:samplingfunction}c.

This example demonstrates that variance-based pruning offers a more effective clustering strategy for TNTRules than traditional distance-based pruning. In the given scenario, the global optimum is enclosed within cluster 4. While distance-based pruning groups a large area into this cluster—overlooking the underlying uncertainty of the GP model—variance-based pruning adapts to the model’s uncertainty. Regions with higher uncertainty, where human intervention might be necessary, are clustered more densely, whereas areas with higher GP certainty form larger, less granular clusters. 

\section{Optimizing for Explanation Quality}\label{sec:optimization}

We formulate the challenge of providing good explanations as an optimization problem. We maximize the quality of explanations by exploiting the controllability property of TNTRules. We argue that explanation quality depends on a trade-off between several metrics. We formulate this as multi-objective optimization. Multi-objective optimization is a process of finding optimal values when multiple competing objectives need to be met. 

We seek an optimal $t_s$ value (cf. Sec~\ref{sssec:clustering}) that simultaneously maximizes support, relevance, and rule set length (equivalent to minimizing the coverage of each rule): $\mathbf{{t_s}_{opt}} = \mathbf{\argmax}_{t_s\in[0,1]} [Supp, Rel, |\boldsymbol{\rho}|]$.  The range between [0,1] is set because we are normalizing the variances in the data to be in [0,1].

We choose these metrics to ensure that the explanations effectively identify solutions (Relevance) and provide accurate localization with valid data support (Support and rule set length). While maximizing the rule set length may seem counterintuitive to having a small explanation set, in our case we filter the rule set in the last step of the algorithm~\ref{tntrules} to keep the final explanations (\textit{FilterRules} (Sec~\ref{sssec:rulerank}), so the overall length of the explanation set remains small. We get a Pareto front of solutions in this case.

We can also frame this as a scalar optimization problem $t_{s_{opt}} = \argmax_{t_s \in [0,1]} \alpha(t_s)$ with interestingness $\alpha$ as the metric. In this case, the choice of weights for $\alpha$ becomes crucial, as it involves balancing the different aspects of all the metrics. Nevertheless, we found the multiobjective optimization method easier to automatically balance the different metrics, at the cost of more time due to multiple runs.

\section{Experimental Setup}\label{sec:evaluation}
This section describes our evaluation metrics and criteria, threshold tuning, and experimental setup. For our experiments, we selected benchmark functions with known ground truth and a common Bayesian Optimization (BO) problem: hyperparameter optimization (HPO) of machine learning models, using publicly available data.~\cite{snoek2012practical}.

\begin{figure*}[!tb]
     \centering
     \begin{subfigure}[b]{0.48\textwidth}
         \centering
          \includegraphics[width=\linewidth]{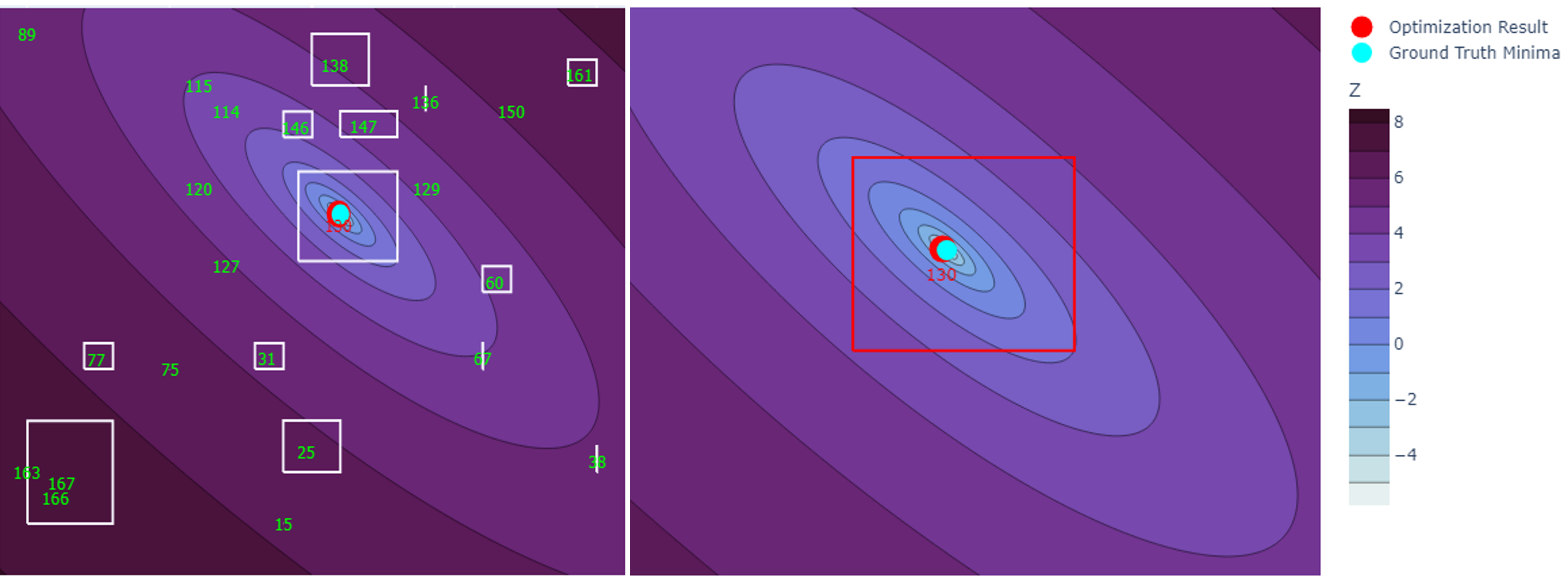}
         \caption{Booth Function}
     \end{subfigure}
     \begin{subfigure}[b]{0.48\textwidth}
         \centering
          \includegraphics[width=\linewidth]{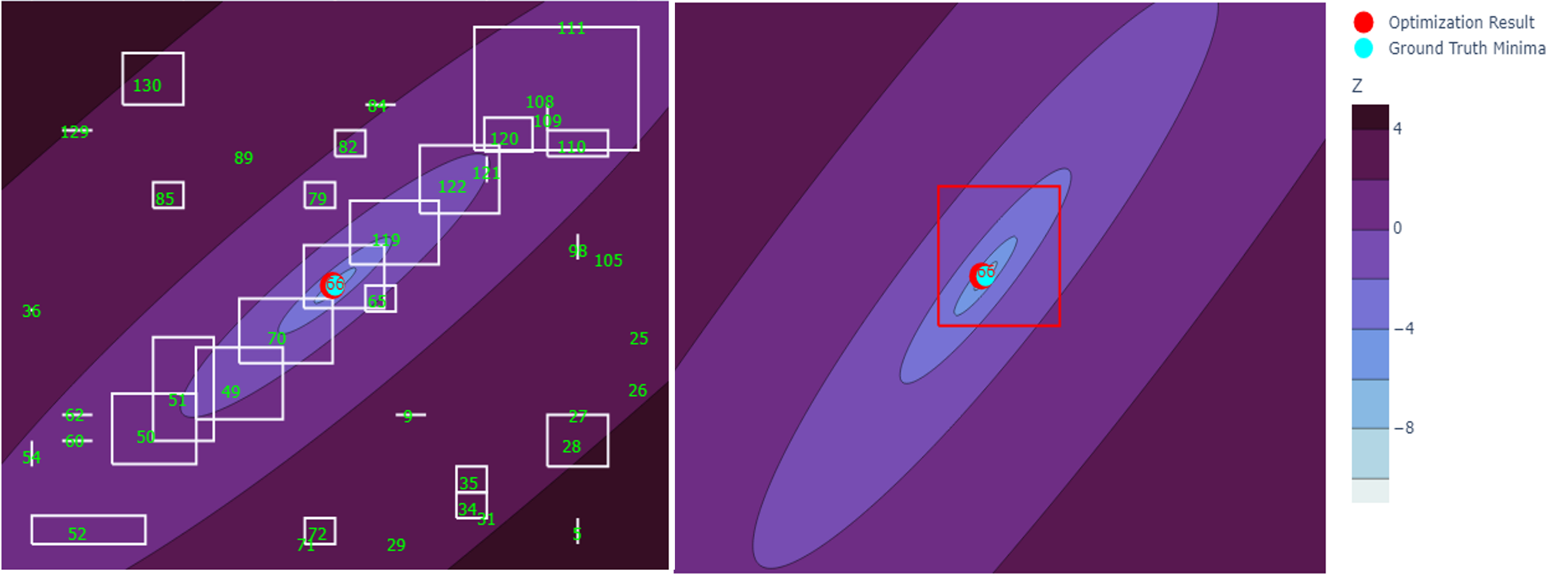}
         \caption{Matyas function}
     \end{subfigure}
     \begin{subfigure}[b]{0.48\textwidth}
         \centering
         \includegraphics[width=\linewidth]{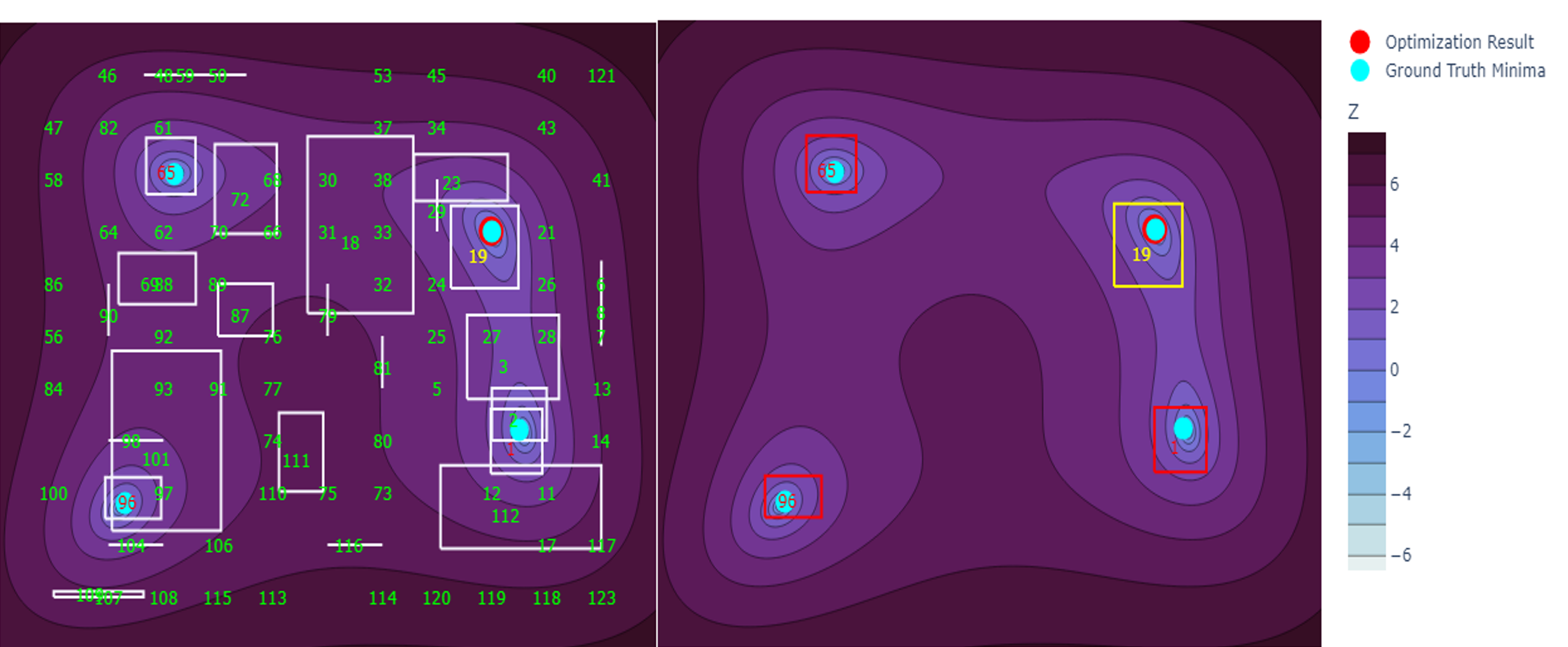}
         \caption{Himmelblau function}
     \end{subfigure}
     \begin{subfigure}[b]{0.48\textwidth}
         \centering
         \includegraphics[width=\linewidth]{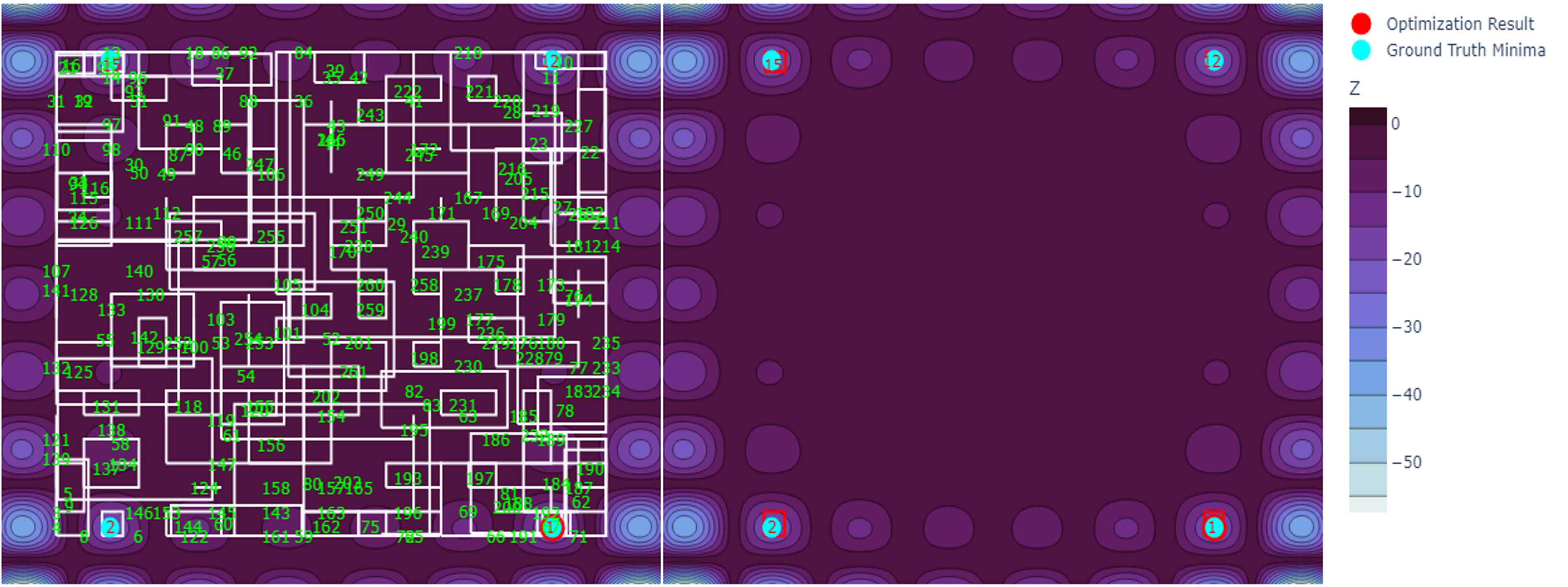}
         \caption{Hölder table function}
     \end{subfigure}
      \begin{subfigure}[b]{0.48\textwidth}
         \centering
          \includegraphics[width=\linewidth]{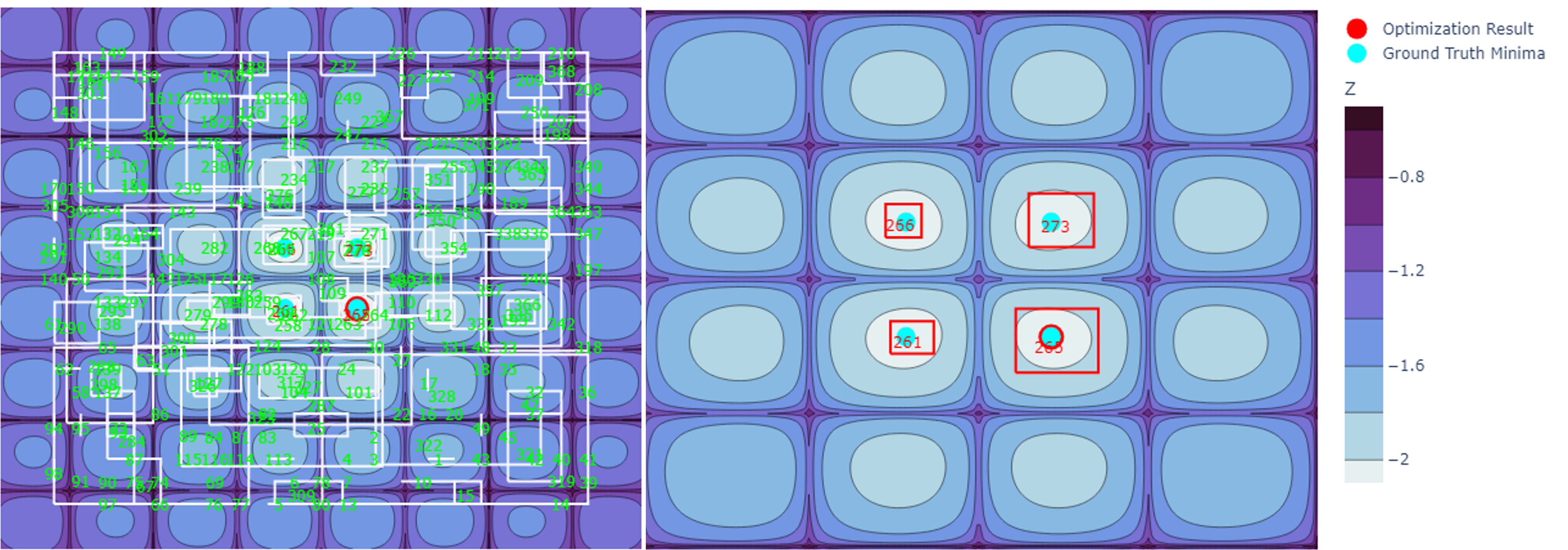}
         \caption{Cross-in-Tray function}
     \end{subfigure}

    \caption{Results of benchmark test functions marked with minima locations, red dot indicates BO found minima. Bounding boxes represent rule-based explanations. White boxes on the left show all rules from the HAC method. The final explanations after filtering are shown on the right. Highly relevant rules are shown in red boxes, moderately relevant rules are shown in yellow boxes, and irrelevant rules are omitted.}
    \label{fig:bofunc}
\end{figure*}

\subsection{Evaluation Criteria and Metrics}\label{ssecmetrics}
We use the three evaluation metrics suitable for our use case from the Co-12 properties to evaluate TNTRules~\cite{10.1145/3583558}: correctness, completeness, and compactness. 

\textbf{Correctness}: Measures the correctness of explanations with respect to the model (also called fidelity). Traditional decision boundary fidelity measures are inappropriate for continuous distributions, so we use a sampling-based approach. Let $\mathbf{N}$ be the number of uniformly sampled points from the rule antecedents. Define $F(x)$ as a function that returns 1 if the GP prediction of $N_i$ is within the bounds of the rule consequent and 0 otherwise. The mean fidelity, $\bar{F}$, is computed as $\bar{F} = \frac{1}{N} \sum_{i=1}^{N} F(x_i)$ where $x_i \in \mathbf{X}$. High correctness ensures that the explanations are reliable and can be trusted by the users. In practical terms, the expert can depend on these suggestions to make adjustments that agree with the learned model and are not based on erroneous or misleading information.

\textbf{Compactness}: Measures the global size of the ruleset ($|\boldsymbol{\rho}|$) (Sec.~\ref{sssec:rulerank}) to determine how large the ruleset is, which is directly related to the effort required to understand it. The practical benefit of compactness is that it makes the explanations accessible and actionable. Users are more likely to utilize insights that are easy to understand. Compact explanations would reduce cognitive load, making it easier for experts to quickly grasp and act on recommendations without extensive analysis.

\textbf{Completeness}: Measures how well input parameters are represented in a rule. It is the ratio of the parameters represented in the longest rule in a set to the total number of input parameters. ($\frac{|\rho_{\dashv}|_{max}}{|p|}$) (Sec.~\ref{sssec:ruleconst}). Completeness ensures that the explanation is completely describing all parameters of the problem. For practical purposes, the explanations should offer recommendations that consider all parameters without ignoring any. This is crucial for ensuring that no significant aspect of the tuning process is ignored, potentially leading to clear explanations.

\vspace{-2mm}
\subsection{Optimizing Clustering Threshold ($t_s$)}\label{ssec:tuningcutoff}
To determine $t_s$, we use the pymoo NSGA II implementation\footnote{\url{https://pymoo.org/algorithms/moo/nsga2.html}} with 25 iterations. We set the population size to 25, the number of generations to 10, and initialized the algorithm with no constraints. We select the solution with the highest rule set length from the Pareto front.  


\subsection{Optimization Problems}\label{ssec:optprob}
We evaluate TNTRules on two sets of optimization problems: First, we benchmark using test functions for optimization (Booth, Matyas, Himmelblau, Hölder table, and Cross-in-tray functions)~\cite{mishra2006some,silagadze2007finding,Himmelblau1972AppliedNP}, and second, HPO with deep learning models (MLP, ResNET~\cite{7780459}, and XceptionNET~\cite{8099678}). We set the weights for interestingness $\alpha$ as $w_1 = 0.2, w_2 = 0.2, w_3 = 0.1, w_4 = 0.5$ (Sec.~\ref{sssec:rulerank}) for all experiments in this paper as recommended by the multi-objective optimization. Ablation for different weights was also explored to show how well the optimization step performs (cf. Sec~\ref{sssec:intweights}). We perform a comparative analysis of TNTRules with the XAI methods Decision trees (DT)~\cite{10.5555/3044805.3044891}, RXBO~\cite{chakraborty_post-hoc_nodate}, and RuleXAI~\cite{macha2022rulexai}.

\subsubsection{Optimization Benchmark Functions}\label{ssec:bofunc}
We evaluate TNTRules on five optimization benchmark functions (Booth, Matyas, Himmelblau, Hölder table, and Cross-in-tray functions). These domains allow the collection of inexpensive baseline samples that allow for direct evaluation. BO samples data from the respective search spaces for the benchmark functions to perform optimization. For the TNTRules explanation set generation, we also use the same search space to sample the explanation set.

\subsubsection{Hyperparameter Optimization Problems}\label{ssec:HPO}

We evaluate the performance of TNTRules in a practical deep-learning context involving HPO tasks for both classification and regression problems. We use the MNIST~\cite{deng2012mnist}, CIFAR-10~\cite{krizhevsky2009learning}, and California Housing~\cite{wu2020prediction} datasets for the study. All datasets were randomly divided into 80-10-10 training, testing, and validation sets. We used Tensorflow and Keras for the experiments. We normalize the data as a pre-processing step for MNIST and CIFAR-10. For California Housing, we one-hot encode the categorical features as a preprocessing step. We also use the standard Keras tuner to optimize the model parameters using BO\footnote{\url{https://www.tensorflow.org/tutorials/keras/keras_tuner}}. We set validation accuracy as the objective for the optimization problem in the Keras tuner. The BO algorithm behind the Keras tuner framework also samples the hyperparameters from a defined search space. For the declarative set in TNTRules, we use the same search space definition as the Keras tuner to perform the sampling. 
In these tasks, the number of minima is unknown beforehand; thus, selecting a threshold is difficult. We create the training data set for TNTRules and use it as input for all methods. The fidelity of the methods is determined by using GP samples since ground truth samples are too expensive to obtain.

For TNTRules, we focus on rules with interestingness $\alpha > 0.7$, i.e. rules that have a higher chance of being an alternative solution. 


\section{Results} \label{sec:results}

\begin{table*} 
    \begin{minipage}[t]{0.9\linewidth}
     \centering   
     \caption{Results comparing XAI methods on ground truth samples vs. GP surrogate. TNTRules  accurately identify minima in the GP surrogate case and outperform other methods in the ground truth case.}
     \scalebox{0.8}{    
    \begin{tabular}{llrrrrrrr}
    \toprule \tabhead
    &  & \multicolumn{2}{c}{Ground Truth (iter = 1000)} & \multicolumn{2}{c}{Surrogate GP ($N_e$ = 1000)} \\
    \cmidrule(lr){3-4} \cmidrule(lr){5-6} \cmidrule(lr){7-8}
    &  & Compactness & Correctness & Compactness & Correctness \\
    &  & $|\boldsymbol{\rho}|\downarrow$ & $\bar{F}$ $\uparrow$ & $|\boldsymbol{\rho}|\downarrow$ & $\bar{F}$ $\uparrow$ \\
    \otoprule
    \multirow{4}{*}{\rotatebox{90}{\textsc{booth}}} & 
     DR & $700\pm0$ & $0.88\pm0.01$ & $1000\pm0$ & $0.92\pm0.01$ \\
    & RXBO & $10\pm1$ & $0.97\pm0.01$ & $5\pm0$ & $0.97\pm0.01$ \\
    & RuleXAI & $16\pm0$ & $0.30\pm0.00$ & $12\pm1$ & $0.42\pm0.01$ \\
    & \textbf{TNTRules} & $\mathbf{3\pm2}$ & $\mathbf{1\pm0.00}$ & $\mathbf{1\pm0}$ & $\mathbf{0.98\pm0.01}$ \\
    \midrule
    \multirow{4}{*}{\rotatebox{90}{\textsc{Matyas}}} & 
    DR & $700\pm0$ & $0.85\pm0.02$ & $1000\pm0$ & $0.90\pm0.02$ \\
    & RXBO & $15\pm0$ & $0.98\pm0.01$ & $6\pm5$ & $0.97\pm0.02$ \\
    & RuleXAI & $17\pm1$ & $0.27\pm0.00$ & $50\pm2$ & $0.48\pm0.02$ \\
    & \textbf{TNTRules} & $\mathbf{6\pm1}$ & $\mathbf{1\pm0.00}$ & $\mathbf{1\pm0}$ & $\mathbf{0.99\pm0.00}$ \\
    \midrule
    \multirow{4}{*}{\rotatebox{90}{\textsc{Himmel.}}} & 
    DR & $700\pm0$ & $0.83\pm0.01$ & $1000\pm0$ & $0.98\pm0.01$ \\
    & RXBO & $30\pm3$ & $0.94\pm0.03$ & $25\pm2$ & $0.94\pm0.01$ \\
    & RuleXAI & $17\pm0$ & $0.34\pm0.01$ & $35\pm0$ & $0.55\pm0.01$ \\
    & \textbf{TNTRules} & $\mathbf{6\pm2}$ & $\mathbf{0.99\pm0.01}$ & $\mathbf{4\pm1}$ & $\mathbf{0.99\pm0.00}$ \\
    \midrule
    \multirow{4}{*}{\rotatebox{90}{\textsc{Holder.}}} & 
    DR & $699\pm1$ & $0.88\pm0.03$ & $1000\pm0$ & $0.77\pm0.03$ \\
    & RXBO & $14\pm1$ & $0.97\pm0.02$ & $12\pm2$ & $\mathbf{0.98\pm0.01}$ \\
    & RuleXAI & $17\pm0$ & $0.78\pm0.00$ & $55\pm1$ & $0.41\pm0.02$ \\
    & \textbf{TNTRules} & $\mathbf{3\pm2}$ & $\mathbf{0.98\pm0.01}$ & $\mathbf{4\pm0}$ & $\mathbf{0.98\pm0.01}$ \\
    \midrule
    \multirow{4}{*}{\rotatebox{90}{\textsc{cross.}}} & 
    DR & $700\pm0$ & $\mathbf{0.99\pm0.00}$ & $1000\pm0$ & $0.97\pm0.02$ \\
    & RXBO & $15\pm5$ & $\mathbf{0.99\pm0.00}$ & $9\pm3$ & $\mathbf{0.99\pm0.00}$ \\
    & RuleXAI & $29\pm1$ & $0.96\pm0.02$ & $127\pm1$ & $\mathbf{0.99\pm0.00}$ \\
    & \textbf{TNTRules} & $\mathbf{4\pm0}$ & $0.98\pm0.00$ & $\mathbf{4\pm0}$ & $\mathbf{0.99\pm0.00}$ \\
    \bottomrule
    \end{tabular}}
    \label{tab:funcres}
    \end{minipage}%
    
    \begin{minipage}{0.9\linewidth}
       \centering
       \caption{Comparison of XAI methods with TNTRules for HPO with BO for deep learning models (MLP, ResNET, and XceptionNET). We utilized MNIST, California housing (Housing), and the CIFAR10 dataset. TNTRules outperform all methods on all three criteria.  }
       \scalebox{0.8}{
    \begin{tabular}{lllrrr}
    \toprule\tabhead
         & &  & Compactness  &   Completeness  & Correctness  \\
         & &  & $|\boldsymbol{\rho}|\downarrow$ & $|\rho_{\dashv}|_{max}/|p| == 1.00$ & $\bar{F}$ $\uparrow$ \\
        \otoprule
\multirow{8}{*}{\rotatebox{90}{\textsc{MLP (p = 6)}}}&
\multirow{4}{*}{\rotatebox{90}{\textsc{MNIST}}}
         & DR     & $141\pm5$    &$4.33\pm0.02$  & $0.02\pm0.01$\\
         & & RXBO & $62\pm5$     &$\mathbf{1.00\pm0.00}$   & $0.80\pm 0.01$  \\ 
         & & RuleXAI & $42\pm1$     &$1.33\pm0.00$   & $0.01\pm0.00$ \\
         & & \textbf{TNTRules} & $\mathbf{23\pm 1}$ & $\mathbf{1.00\pm0.00}$   & $\mathbf{0.85\pm0.01}$ \\\cmidrule{3-6}
& \multirow{4}{*}{\rotatebox{90}{\textsc{Housing}}} 
        & DR      & $154\pm6$ &  $2.50\pm0.04$   & $0.01\pm0.00$ \\
        & & RXBO  & $34\pm8$ & $\mathbf{1.00\pm0.00}$ & $0.61\pm0.02$ \\ 
        & & RuleXAI & $34\pm1$ & $1.50\pm0.00$ & $0.01\pm0.00$\\
        & & \textbf{TNTRules}  & $\mathbf{22\pm1}$ & $\mathbf{1.00\pm0.00}$ & $\mathbf{0.82\pm0.03}$ \\\midrule
\multirow{8}{*}{\rotatebox{90}{\textsc{ResNet (p = 6)}}}&
\multirow{4}{*}{\rotatebox{90}{\textsc{MNIST}}}
        & DR      & $1000\pm15$ & $2.83\pm0.02$ & $0.93\pm0.01$ \\
        & & RXBO  & $55\pm2$ & $\mathbf{1.00\pm0.00}$ & $0.97\pm0.01$ \\ 
        & & RuleXAI & $117\pm0$ & $1.50\pm0.00$ & $0.86\pm0.02$ \\
        & & \textbf{TNTRules} & $\mathbf{3\pm0}$ & $\mathbf{1.00\pm0.00}$ & $\mathbf{1\pm0.00}$\\\cmidrule{3-6}
& \multirow{4}{*}{\rotatebox{90}{\textsc{CIFAR10}}} 
        & DR      & $1000\pm13$ & $1.66\pm0.00$ & $0.98\pm0.01$ \\
        & & RXBO  & $57\pm5$ & $\mathbf{1.00\pm0.00}$ & $0.99\pm0.00$\\ 
        & & RuleXAI & $106\pm0$ & $2.00\pm0.00$ & $0.72\pm0.01$\\
        & & \textbf{TNTRules} & $\mathbf{12\pm0}$ & $\mathbf{1.00\pm0.00}$ & $\mathbf{1\pm0.00}$ \\\midrule
\multirow{6}{*}{\rotatebox{90}{\textsc{\small{Xcep.NET (p = 11)}}}}&
\multirow{4}{*}{\rotatebox{90}{\textsc{MNIST}}}
        & DR     & $56\pm2$ & $0.54\pm0.00$ & $0.98\pm0.01$ \\
        & & RXBO & $69\pm2$ & $\mathbf{1.00\pm0.00}$ & $0.99\pm0.00$ \\ 
        & & RuleXAI & $141\pm1$ & $0.72\pm0.02$ & $0.99\pm0.00$ \\
        & & \textbf{TNTRules} & $\mathbf{4\pm0}$ & $\mathbf{1.00\pm0.00}$ & $\mathbf{1\pm0.00}$\\\cmidrule{3-6}
& \multirow{4}{*}{\rotatebox{90}{\textsc{CIFAR10}}} 
        & DR    & $977\pm3$ & $2.27\pm0.02$ & $\mathbf{0.84\pm0.02}$ \\
        & & RXBO & $21\pm3$ & $\mathbf{1.00\pm0.00}$ & $0.80\pm0.01$\\ 
        & & RuleXAI & $130\pm0$ & $0.72\pm0.00$ & $0.14\pm0.01$ \\
        & & \textbf{TNTRules} & $\mathbf{5\pm1}$ & $\mathbf{1.00\pm0.00}$ & $0.81\pm0.05$\\
        \midrule
    \end{tabular}}
     \label{tab:hporesults}
    \end{minipage}
\end{table*}

In this section, we first present qualitative and quantitative results on the optimization benchmark functions (Sec.~\ref{sec:results:qualitative-benchmark}), confirming that TNTRules extracts relevant regions of interest and significantly reduces the search space. We then present the quantitative evaluation for the HPO use case (Sec.~\ref{sec:results:benchmark}), showing the superiority of TNTRules over the state-of-the-art XAI baselines. We then show through an ablation study that the GP approximation is a valid way to evaluate TNTRules (Sec.~\ref{sssec:abs}). Finally, we ablate for clustering choices and pruning (Sec.~\ref{ssec:ablation}).

\begin{table}
\caption{Summary of results (Tab.~\ref{tab:funcres} and~\ref{tab:hporesults}) explainability methods mapped to evaluation criteria, TNTRules outperforms baselines.}
    \centering\scalebox{0.8}{
    \begin{tabular}{+l^c^c^c^c}
    \toprule \tabhead
          Methods & Compactness & Completeness & Correctness  \\
        \otoprule
         Decision Rules & High number of rules & No & Low\\
         RXBO & Moderate number of rules & \textbf{Yes} & \textbf{High}\\
         RuleXAI & Moderate number of rules & No & Low\\
        \textbf{TNTRules} & \textbf{Low number of rules} & \textbf{Yes} & \textbf{High} \\
       \bottomrule
    \end{tabular}}
    \label{tab:condensedresults}
\end{table}

\subsection{Results on Benchmark Functions}\label{sec:results:qualitative-benchmark}
Fig.~\ref{fig:bofunc} shows the resulting explanations for the benchmark functions visualized in function space. The left hand side of each pair of plots shows all detected rules $\boldsymbol{\rho_{all}}$  before filtering, with white boxes indicating their boundaries. The right hand side, shows the rules after filtering: red boxes ($\alpha \geq 0.6$) indicate highly interesting rules , yellow boxes ($0.6 > \alpha \geq 0.4$) indicated moderately interesting rules. Rules with lower interest levels are omitted. 
\textbf{\textit{The results show that TNTRules effectively localized high-interest minima identified by BO in four out of five cases}} except for the Himmelblau's function where the generated rule showed moderate interestingness. The rule selection process reduced the number of initially discovered rules by 98.5\%, highlighting key rules as explanations. Rule curation resulted in a significant 98\% reduction in search space exploration when BO was rerun.

\vspace{-2mm}

\subsection{Results on Hyperparameter Optimization Problems}\label{sec:results:benchmark}
\label{ssec:results:hpo}
The performance comparison in Tab.~\ref{tab:hporesults} shows that, \textbf{\textit{explanations generated by TNTRules outperform all the baseline methods in all metrics.}} Both, TNTRules and RXBO, consistently demonstrate high correctness in their explanations, reflecting a robust approximation of the underlying GP model, which is facilitated by the clustering-based surrogate. 

Analysing results in more detail, we observe that the decision rules tend to overfit the data, resulting in an excessive number of rules (compactness criterion). In contrast, RuleXAI and RXBO struggle to fit the data adequately, as evidenced by the moderate number of rules they generate. The variance-based pruning method introduced in TNTRules consistently achieves better cluster separations, leading to a reduced number of rules and a more compact rule set. This trend aligns with the findings from the ablation study using test functions.

Regarding completeness, we see that methods such as DT and RuleXAI produce rules of variable length. These rules may either repeat parameters across different ranges, increasing overall rule length, or omit parameters completely, resulting in shorter rules. In our case, completeness is greater than 1 for longer rules and less than 1 for shorter ones. In contrast, TNTRules by design creates fixed-length rules that correspond to the number of input parameters. By neither omitting nor repeating parameters, we ensure completeness remains at 1. This comprehensive representation of parameters within the rules eliminates confusion for engineers tuning physical systems, as they require clarity on the fixed ranges of each parameter. Overall, the results indicate that TNTRules outperforms other XAI methods both qualitatively and quantitatively.

An overall qualitative summary of the results is shown in Tab.~\ref{tab:condensedresults}. 

\vspace{-2mm}
\subsection{Ablation for Interestingness Weights}\label{sssec:intweights}
\begin{table*}
       \centering
       \caption{Ablation showing the impact of optimized interestingness weights on TNTRules. We test five different weight combinations for TNTRules besides the optimal version and compare them with baseline RXBO to show the impact on metrics compactness and correctness. Results show optimized weights TNTRules outperform the rest, thus highlighting the importance of setting the correct weights.  }
       \scalebox{0.7}{
    \begin{tabular}{lllrrr}
    \toprule\tabhead
         & &  & Configuration  &   Compactness  & Correctness  \\
         & &  & $w_1, w_2, w_3, w_4$ & $|\boldsymbol{\rho}|\downarrow$ & $\bar{F}$ $\uparrow$ \\
        \otoprule
\multirow{14}{*}{\rotatebox{90}{\textsc{MLP (p = 6)}}}&
\multirow{7}{*}{\rotatebox{90}{\textsc{Housing}}}
         & RXBO  & - & $34\pm8$ & $0.61\pm0.02$ \\ 
         & & TNTRules & $w_1 = 0.1, w_2 = 0.1, w_3 = 0.1, w_4 = 0.7$     &$56\pm1$   & $0.76\pm 0.01$  \\ 
         & & TNTRules & $w_1 = 0.1, w_2 = 0.2, w_3 = 0.2, w_4 = 0.5$     &$32\pm2$   & $0.77\pm 0.02$  \\
         & & TNTRules & $w_1 = 0.2, w_2 = 0.1, w_3 = 0.2, w_4 = 0.5$     &$30\pm4$   & $0.72\pm 0.03$  \\
         & & TNTRules & $w_1 = 0.2, w_2 = 0.1, w_3 = 0.1, w_4 = 0.6$     &$45\pm1$   & $0.78\pm 0.01$  \\
         & & TNTRules & $w_1 = 0.1, w_2 = 0.1, w_3 = 0.3, w_4 = 0.5$     &$47\pm4$   & $0.71\pm 0.04$  \\
         & & \textbf{TNTRules w Optimization} & $w_1 = 0.2, w_2 = 0.2, w_3 = 0.1, w_4 = 0.5$ & $\mathbf{23\pm 1}$   & $\mathbf{0.85\pm0.01}$ \\\cmidrule{3-6}
& \multirow{7}{*}{\rotatebox{90}{\textsc{MNIST}}} 
        &  RXBO &    -  & $62\pm5$   & $0.80\pm 0.01$  \\ 
        & & TNTRules & $w_1 = 0.1, w_2 = 0.1, w_3 = 0.1, w_4 = 0.7$     &$46\pm3$   & $0.81\pm 0.02$  \\ 
        & & TNTRules & $w_1 = 0.1, w_2 = 0.2, w_3 = 0.2, w_4 = 0.5$     &$26\pm1$   & $0.81\pm 0.01$  \\
        & & TNTRules & $w_1 = 0.2, w_2 = 0.1, w_3 = 0.2, w_4 = 0.5$     &$26\pm2$   & $0.77\pm 0.03$  \\
        & & TNTRules & $w_1 = 0.2, w_2 = 0.1, w_3 = 0.1, w_4 = 0.6$     &$31\pm2$   & $0.82\pm 0.01$  \\
        & & TNTRules & $w_1 = 0.1, w_2 = 0.1, w_3 = 0.3, w_4 = 0.5$     &$39\pm1$   & $0.80\pm 0.01$  \\
        & & \textbf{TNTRules w Optimization}  & $w_1 = 0.2, w_2 = 0.2, w_3 = 0.1, w_4 = 0.5$ & $\mathbf{22\pm1}$ & $\mathbf{0.82\pm0.03}$ \\\midrule
        \midrule
    \end{tabular}}
     \label{tab:abhporesults}
\end{table*}
Interestingness is an important factor in TNTRules to generate good explanations. To recall, we defined interestingness of a rule as the weighted sum of Coverage (Covr), Support (Supp), Confidende (Con), and Relevance (Rel): $\alpha = w_1*Covr + w_2*Supp + w_3*Con + w_4*Rel$ (cf. Sec.~\ref{sssec:rulerank}).

In previous sections, the weights for interestingness, $\alpha$ were determined through a multi-objective optimization process, resulting in the values $w_1 = 0.2, w_2 = 0.2, w_3 = 0.1, w_4 = 0.5$. In this ablation study, we investigate the impact of variations in these weights on the compactness and correctness of explanations. We omit completeness, as TNTRules explanations are complete by design (cf. Sec.~\ref{ssec:results:hpo}).

We investigated how weight changes influence our quality metrics. We evaluate on two datasets: MNIST for image classification and California housing for regression tasks using the MLP model. TNTRules was applied to the HPO task using BO on the MLP model for both the dataset. We tested five different configurations of the weights $w_1, w_2, w_3, w_4$, each chosen randomly close to the optimized weights. 

The results (Tab.~\ref{tab:abhporesults}) indicate that \textbf{\textit{ TNTRules configured with optimized weights consistently outperforms both the non-optimized versions of TNTRules and the baseline method, RXBO.}} The non-optimized versions shows a moderate to low number of rules, compromising the compactness of the explanations. This lack of compactness implies that users are required to analyze more extensive information for tuning, potentially increasing cognitive load. Additionally, the correctness of explanations in the non-optimized configurations was lower, often resulting in rules that poorly represent the parameter space. This issue was primarily due to inadequate data support, leading to overly narrow bounding boxes that fail to cover meaningful regions of the parameter space.

The lower quality of rules in the non-optimized configurations could mislead users, providing suboptimal guidance for system tuning. Thus it is important to carefully calibrate the weights. 
\vspace{-2mm}
\subsection{Ablation for GP Approximation}\label{sssec:abs}
So far, we have assumed that the surrogate GP in BO sufficiently approximates the direct BO evaluation of a function (ground truth (GT)) (cf. Sec.~\ref{sec:background}). Thus, we can evaluate on GP samples instead of GT samples, which is useful when obtaining GT samples is costly, as in HPO. Since evaluating optimization test functions is cheap, we perform an ablation study to demonstrate the consistency and independence of our results, regardless of whether we used GT directly or the surrogate GP with an explanatory dataset.

We compared TNTRules with XAI methods under two settings: one with 1000 GT samples ($N_e = 0$) and the other with the sampled explanation dataset and the GP surrogate ($N_e = 1000$). For the fidelity calculation in the GT case, we withheld 25\% of the GT samples. In the GP-based evaluation, fidelity was computed using randomly generated 300 samples, as introduced in Sec.~\ref{ssecmetrics}.

Tab.~\ref{tab:funcres} presents results comparing the GT samples and the GP surrogate with an explanatory dataset. \textbf{\textit{TNTRules outperforms all other methods in the comparison, identifying all minima with high fidelity.}} RXBO performs second best, with better fidelity than decision rules and RuleXAI, suggesting good data approximation. Decision rules overfit, as evidenced by numerous rules, while RuleXAI underfits, as evidenced by low fidelity scores. An exception is the Cross-in-Tray function, where all methods had high fidelity. 

In case the GP is not able to sufficiently approximate the black box function (indicating a bad BO run), our explanations are still faithful to the GP model. The GP model would show high uncertainty, which would also be highlighted in our explanations (through wide ranges of consequences). This visualization of uncertainty can be used as an indication of the quality of the BO run.  

\begin{figure*}[htb]
     \centering
     \begin{minipage}[b]{0.7\textwidth}
         \centering
         \includegraphics[width=\linewidth]{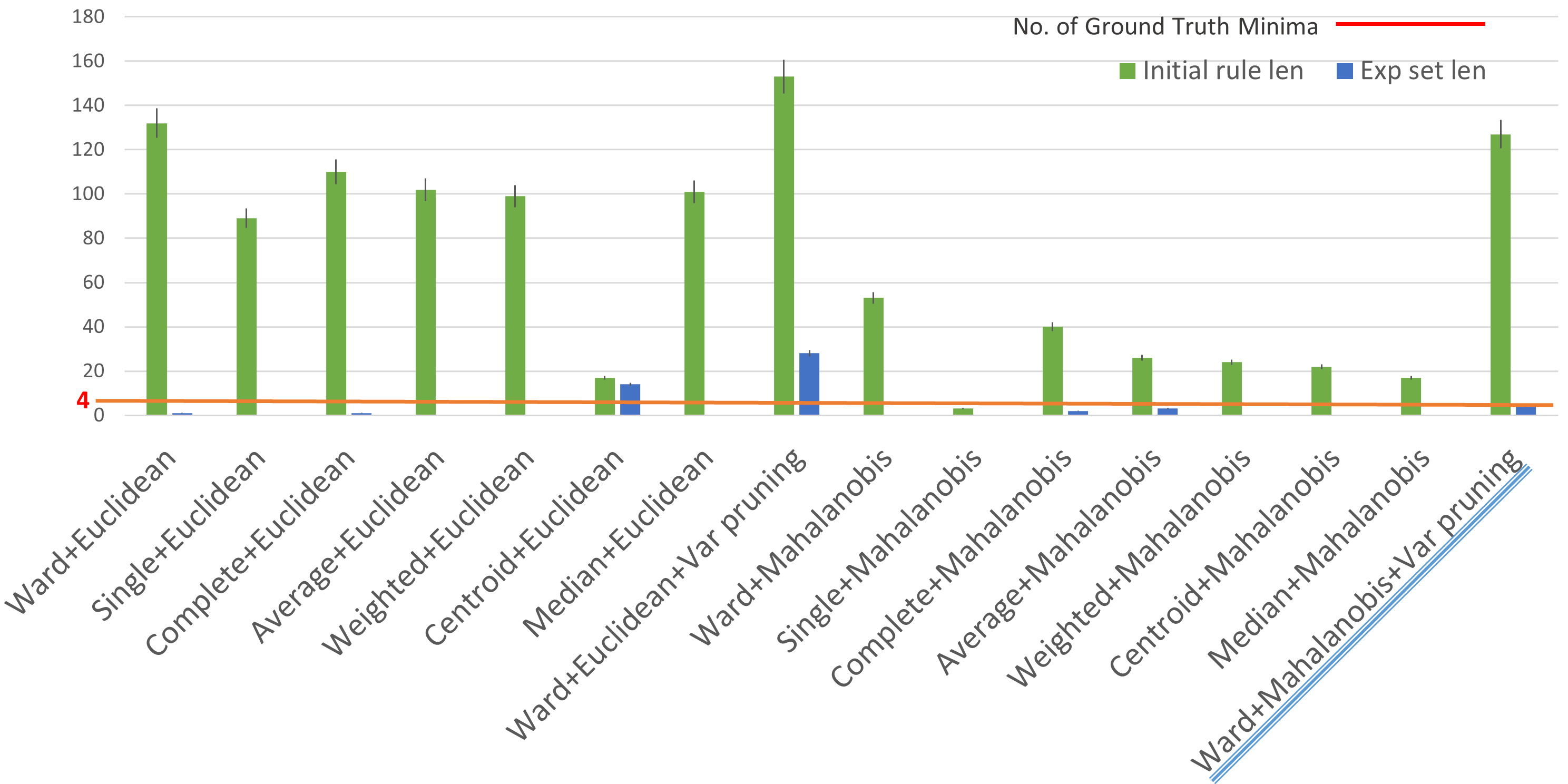}
     \end{minipage}\hspace{1em}
     \begin{minipage}[b]{0.7\textwidth}
         \centering
         \includegraphics[width=\linewidth]{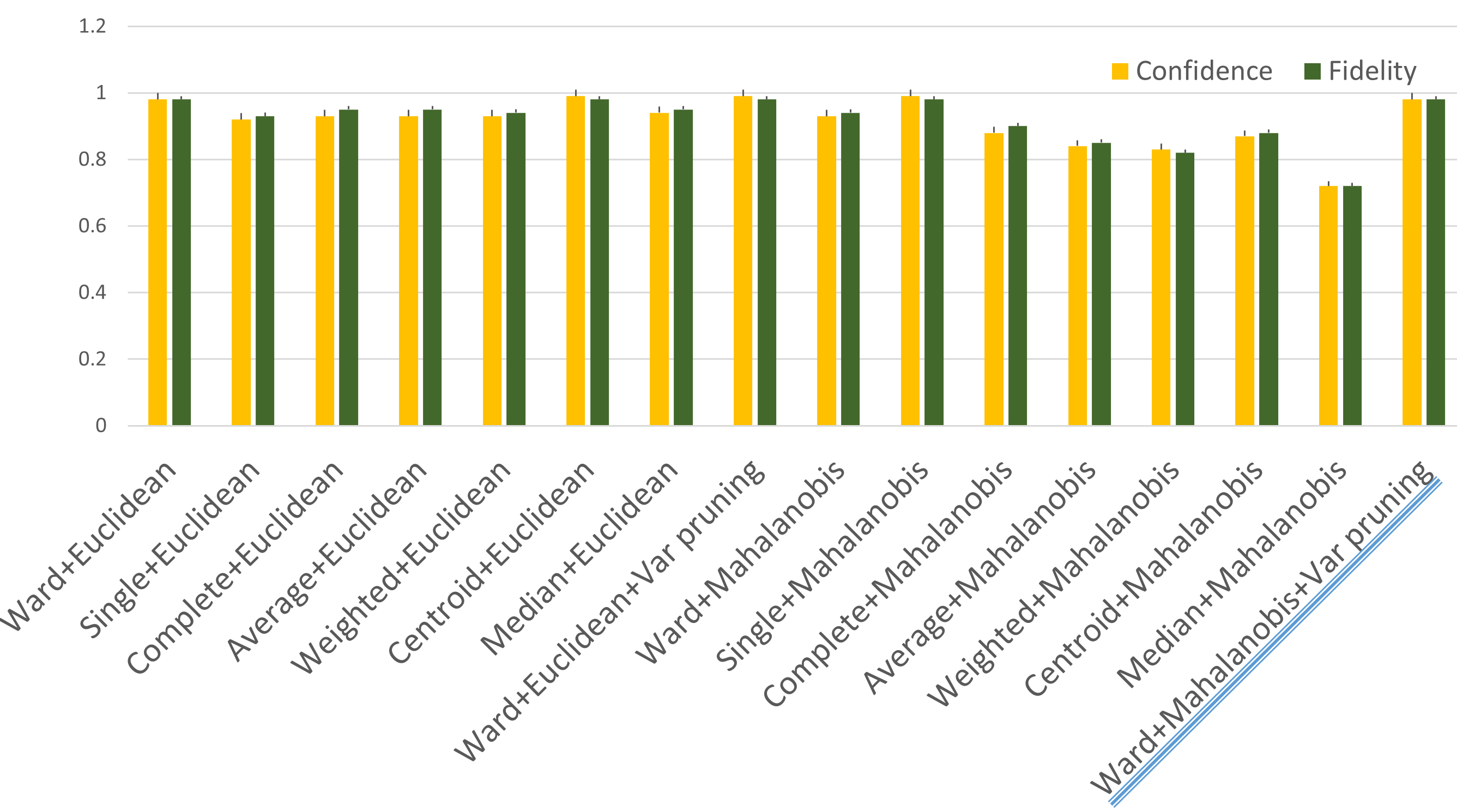}
     \end{minipage}
    \caption{Ablation study comparing 16 HAC configurations. Left: number of initial rules discovered by each method compared to the resulting explanation set after applying the interestingness filter ($\alpha > 0.6$). Right: rule set confidence and fidelity for each configuration. The optimal method is underlined.}
    \label{fig:ablation}
\end{figure*}
\vspace{-2mm}
\subsection{Ablation Clustering Choices and Pruning}\label{ssec:ablation}
We conducted an ablation study to support our design choices for HAC. We compared 16 combinations with seven HAC methods, two distance metrics, distance pruning, and variance pruning. We chose Himmelblau function with four known minima for benchmarking~\cite{Himmelblau1972AppliedNP}. The study evaluated the performance of the methods in identifying and localizing the four minima. The HAC methods included Complete, Ward, Average, Median, Weighted, Centroid, and Single linkages, with Euclidean and Mahalanobis distances~\cite{murtagh2012algorithms}. Each linkage method uses different criteria for clustering the data. Evaluation criteria included the number of initial rules, subset after applying the interestingness filter ($\alpha > 0.6$), correctness, and overall confidence, as shown in Fig.~\ref{fig:ablation}.

From our ablation study, \textbf{\textit{we note the effectiveness of using Ward linkage based on Mahalanobis distance and variance pruning, which accurately identifies all four solutions and provides tight bounding boxes.}} Ward and Complete linkages with Euclidean distance, along with distance-based pruning, yield one highly interesting rule ($\alpha > 0.6$) for one minima and three moderately interesting rules ($0.6 > \alpha \geq 0.4$) covering the remaining three minima. However, the dynamic threshold is challenging when the number of minima is unknown. On the other hand, average linkage with Mahalanobis distance successfully identifies high-interest minima, but results in larger bounding boxes, decreasing the explanatory value, as evidenced by fewer rules found.

\begin{figure*}[!tb]
    \centering
    \begin{subfigure}[b]{0.48\textwidth}
        \centering
        \includegraphics[width=\linewidth]{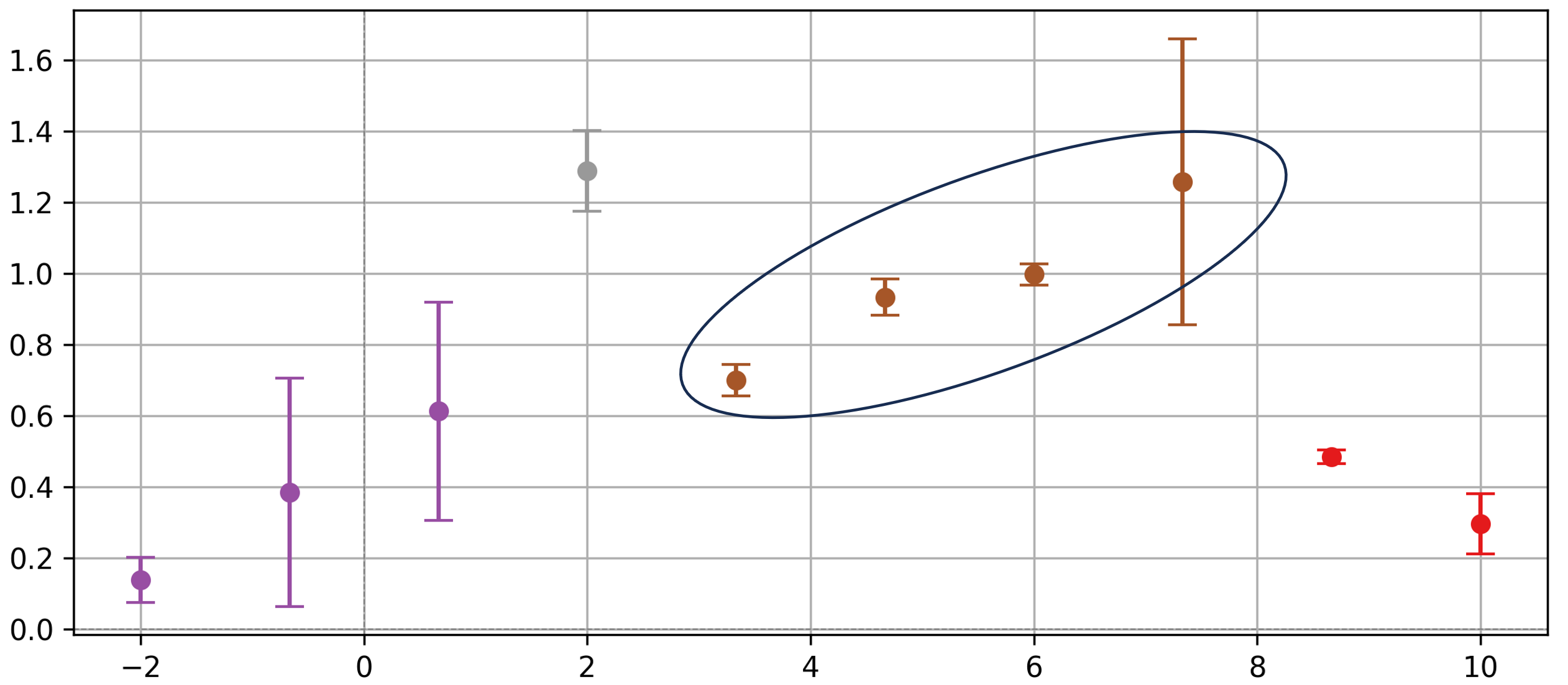}
        \caption{Distance = 1.5}
    \end{subfigure}
    \begin{subfigure}[b]{0.48\textwidth}
        \centering
        \includegraphics[width=\linewidth]{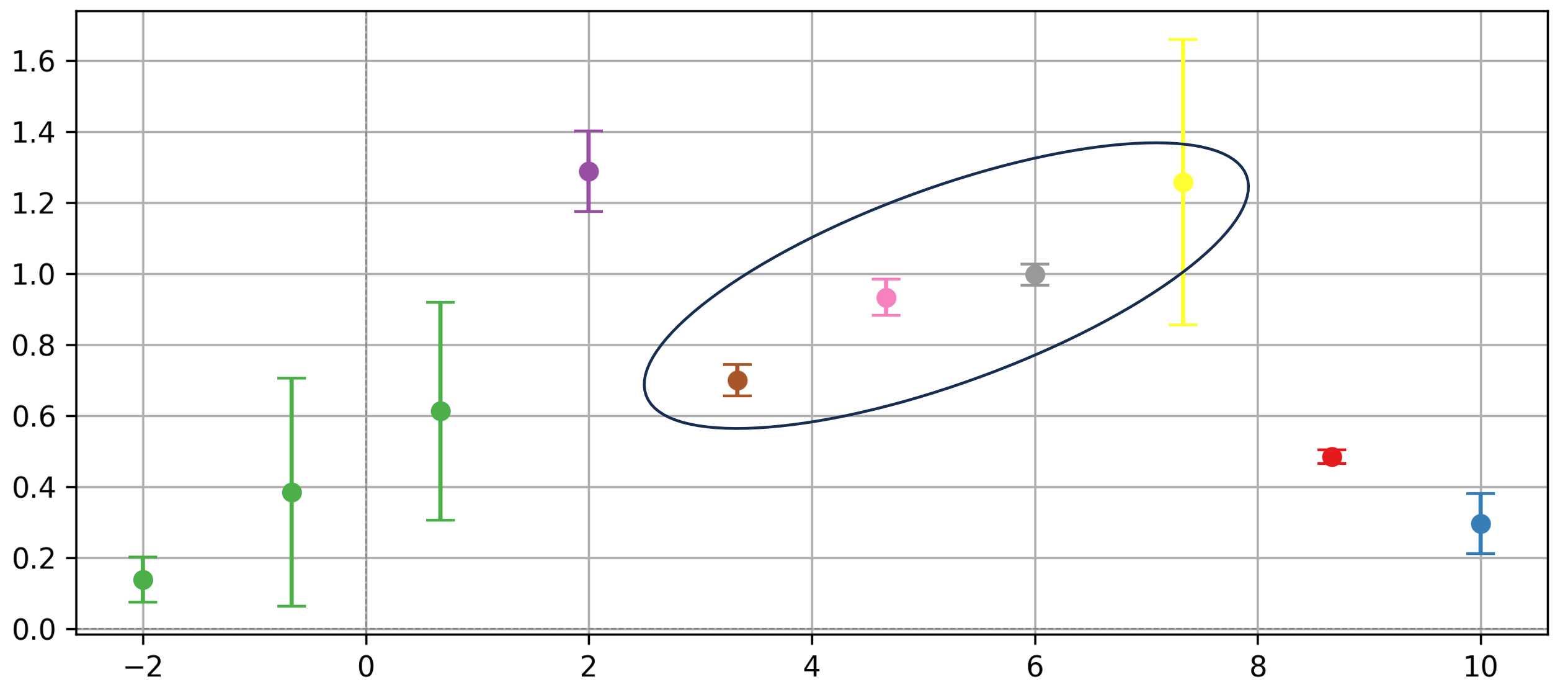}
        \caption{Variance = 1}
    \end{subfigure}
    \begin{subfigure}[b]{0.48\textwidth}
        \centering
        \includegraphics[width=\linewidth]{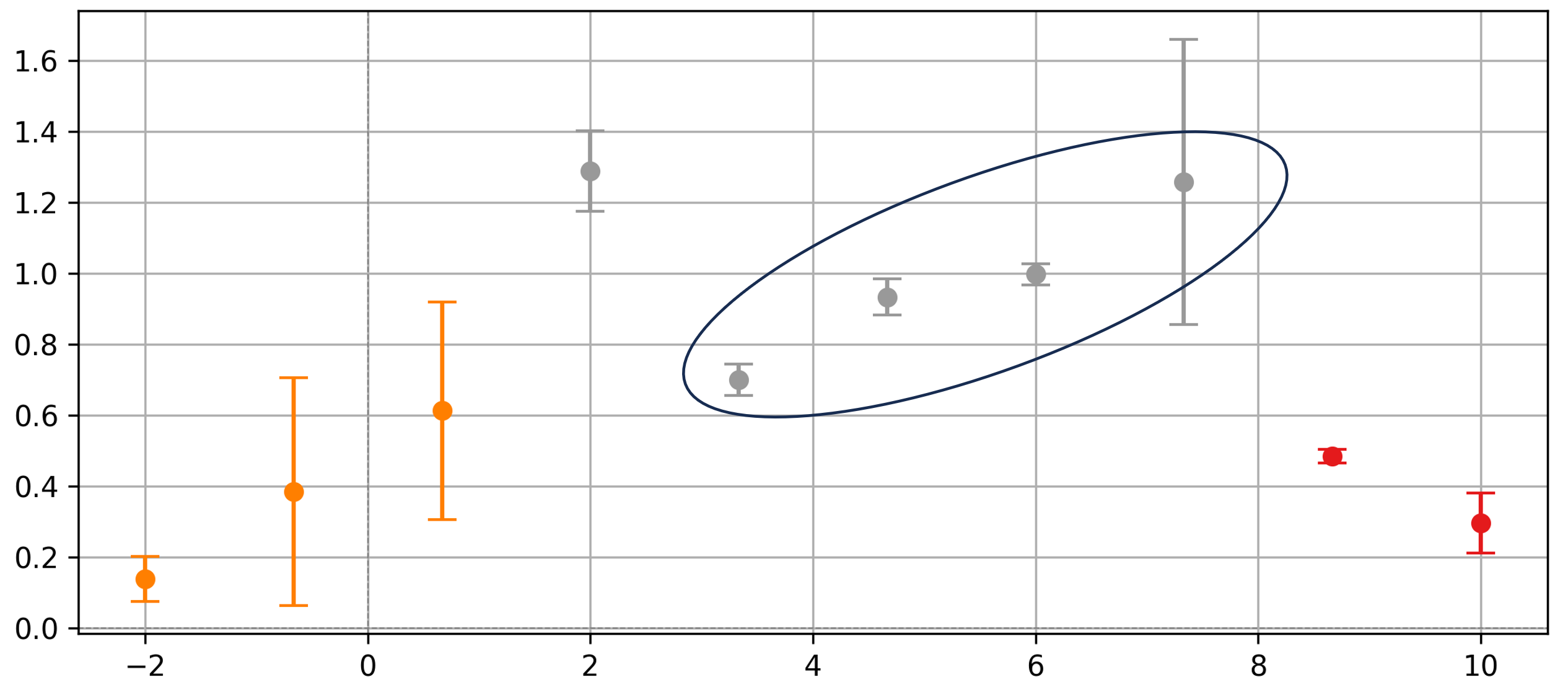}
        \caption{Distance = 1.8}
    \end{subfigure}
    \begin{subfigure}[b]{0.48\textwidth}
        \centering
        \includegraphics[width=\linewidth]{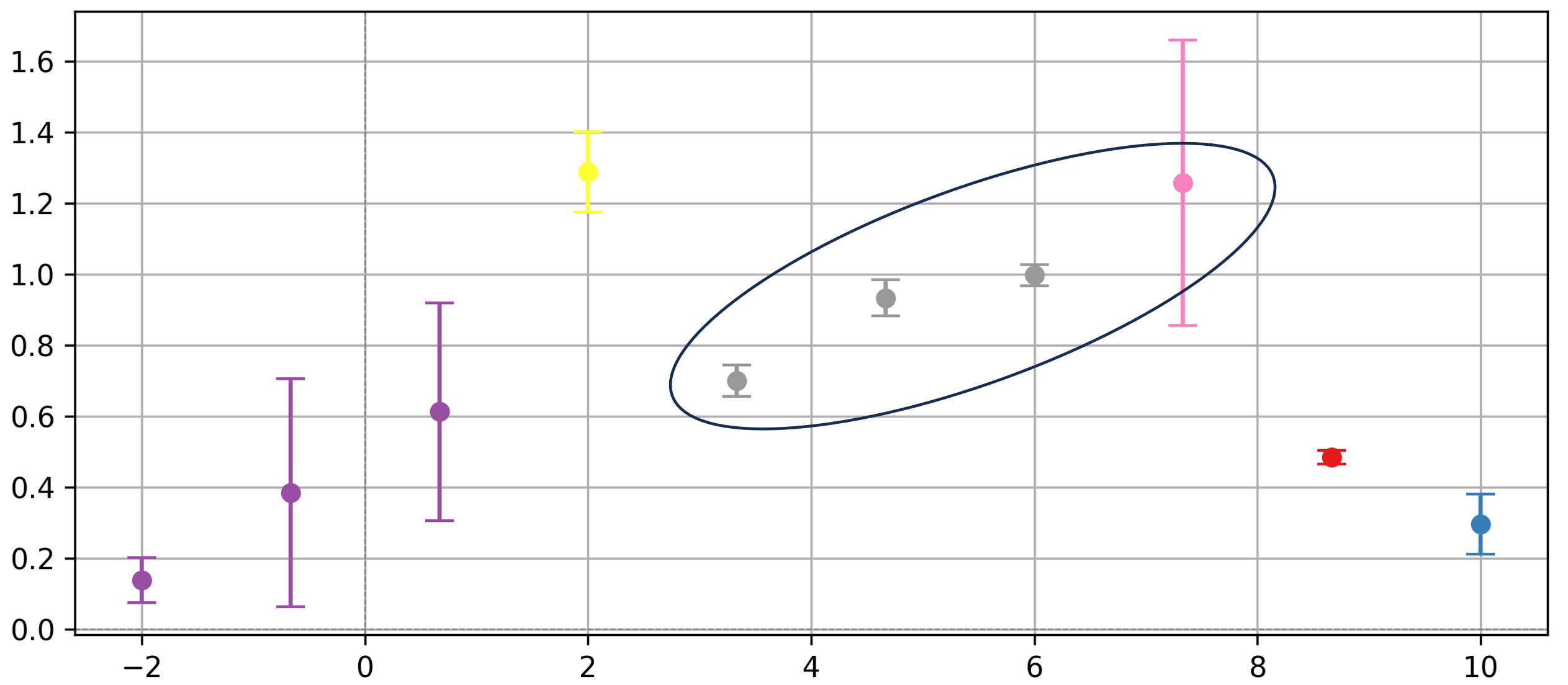}
        \caption{Variance = 5}
    \end{subfigure}
    \caption{Ablation study analyzing the difference in clustering based on different thresholds for distance-based pruning and variance-based pruning. Results in circular highlights show that the distance-based pruning method was not able to separate clusters based on uncertainty, while the variance-based method was able to separate clusters based on uncertainty. }
    \label{fig:absprun}
\end{figure*}

Further, we conducted another ablation study to evaluate our ability to generate uncertainty-aware clusters within the BO framework. To achieve this, we compared standard distance-based pruning using Mahalanobis distance with Ward linkage to our proposed variance-based pruning method. Our analysis used a synthetic problem where we artificially added noise to increase the variance of certain data points. The problem is given as $f(x) = e^{-(x-2)^2} + e^{-(x-6)^2/10} + \frac{1}{x^2 + 1} + \mathcal{N}(0,1)$ This strategy allowed us to assess whether the methods could accurately identify these high variance points and separate them into distinct clusters. 

We used the dendrogram to select threshold points for the distance-based method. For variance pruning, we computed the variance of the subtrees from the linkage matrix and selected different levels of variance as thresholds.  

Our results (Fig.~\ref{fig:absprun}) show that the \textbf{\textit{ distance-based method failed to account for uncertainty, resulting in inadequate cluster separation}}. And this was the expected result since the distance method does not exploit uncertainty~\cite{KLUTCHNIKOFF2022105075}. \textbf{\textit{The variance-based method successfully differentiates clusters based on uncertainty.}} The increase in clusters is not a problem for our use case, because TNTRules filters out clusters with insufficient data support and ranks them based on their interestingness to produce the final explanations. 

\vspace{-2mm}
\section{Limitations and Future Work}\label{sec:limitations}

Even though TNTRules is an effective explainability method for BO, certain aspects warrant further investigation.
First, the effectiveness of TNTRules is based on the performance of the chosen clustering method. While the fidelity of the explanations is preserved through the data support metric for rule selection, the design choices in clustering can significantly impact the range of rules generated. Investigating alternative approaches for grouping data and establishing dynamic criteria for identifying ``good" regions is a promising area of future work.

Second, HAC methods such as clustering with the Ward criterion often struggle with high-dimensional data, leading to challenges in forming meaningful clusters~\cite{10.1007/978-3-030-01851-1_23}. This limitation can adversely affect the performance of TNTRules when applied to very high-dimensional datasets. Standard mitigation techniques such as dimensionality reduction~\cite{VELLIANGIRI2019104}, and approximate clustering~\cite{10066202} can be applied for mitigation. More scenarios need to be explored in the future for generating accurate clusters with high-dimensional data. In addition, if the GP in BO is not sufficiently approximated, TNTRules may produce explanations that are accurate for the model, but not particularly useful to users because they do not accurately reflect the underlying function. Future research could improve the explanations to indicate the degree of GP approximation based on overall uncertainty, thus helping users decide whether to rely on the explanations.

Finally, TNTRules, being a general framework, cover broad industrial application domains; while it is desirable to explore the performance in different domains, we are limited by datasets and access to such industrial applications. Future research could expand upon this by exploring more practical use-cases where TNTRules can be applied.

\vspace{-2mm}
\section{Conclusion}\label{sec:conclusion}
We have introduced and evaluated TNTRules, a post-hoc explainable Bayesian optimization framework designed for collaborative parameter tuning in hardware-centric domains. TNTRules excels at generating high-fidelity, localized rules within the optimization search space, effectively capturing the behavior of the underlying Gaussian Process (GP) model in Bayesian Optimization (BO). This capability allows it to identify alternative solutions that complement traditional BO results, thereby providing actionable tuning recommendations. By delivering comprehensible rules and visualizations, TNTRules enhances the transparency and interpretability of human-AI collaborative tuning in cyber-physical systems. Our experimental results demonstrate that TNTRules surpasses other XAI methods in producing high-quality rules for optimization. In industrial settings, TNTRules could significantly improve the transparency and user-friendliness of automated tuning systems, thus aiding the broader adoption of human-in-the-loop tuning approaches. Future work will involve a user study with a real-world use case to further assess the actionable and human-centric aspects of the explanation formats introduced by TNTRules.

\vspace{-2mm}
\section*{Acknowledgments}
\vspace{-2mm}
This study was supported by BMBF Project hKI-Chemie: humancentric AI for the chemical industry, FKZ 01|S21023D, FKZ 01|S21023G and Continental AG.



 \bibliographystyle{splncs04}

\bibliography{reference}

\begin{thebibliography}{10}
\providecommand{\url}[1]{\texttt{#1}}
\providecommand{\urlprefix}{URL }
\providecommand{\doi}[1]{https://doi.org/#1}

\bibitem{adachi2024looping}
Adachi, M., Planden, B., Howey, D.A., Osborne, M.A., Orbell, S., Ares, N., Muandet, K., Chau, S.L.: Looping in the human collaborative and explainable bayesian optimization (2024)

\bibitem{i2022consistent}
Amoukou, S.I., Brunel, N.J.: Consistent sufficient explanations and minimal local rules for explaining the decision of any classifier or regressor. In: Advances in Neural Information Processing Systems, {NeurIPS}. Curran Associates, Inc. (2022)

\bibitem{apte1997data}
Apt{\'e}, C., Weiss, S.: Data mining with decision trees and decision rules. Future generation computer systems  \textbf{13}(2-3),  197--210 (1997)

\bibitem{chakraborty_post-hoc_nodate}
Chakraborty, T., Wirth, C., Seifert, C.: Post-hoc rule based explanations for black box bayesian optimization. In: Artificial Intelligence. ECAI 2023 International Workshops. pp. 320--337. Springer Nature Switzerland, Cham (2024)

\bibitem{8099678}
Chollet, F.: Xception: Deep learning with depthwise separable convolutions. In: 2017 IEEE Conference on Computer Vision and Pattern Recognition (CVPR). pp. 1800--1807 (2017). \doi{10.1109/CVPR.2017.195}

\bibitem{colliandre2023bayesian}
Colliandre, L., Muller, C.: Bayesian optimization in drug discovery. In: High Performance Computing for Drug Discovery and Biomedicine, pp. 101--136. Springer (2023)

\bibitem{coppens2019distilling}
Coppens, Y., Efthymiadis, K., Lenaerts, T., Now{\'e}, A., Miller, T., Weber, R., Magazzeni, D.: Distilling deep reinforcement learning policies in soft decision trees. In: Proceedings of the workshop on explainable artificial intelligence, IJCAI Workshop. pp.~1--6. IJCAI (2019)

\bibitem{craven1995extracting}
Craven, M., Shavlik, J.: Extracting tree-structured representations of trained networks. Advances in neural information processing systems  \textbf{8} (1995)

\bibitem{deng2012mnist}
Deng, L.: The mnist database of handwritten digit images for machine learning research. IEEE Signal Processing Magazine  \textbf{29}(6),  141--142 (2012)

\bibitem{10.1007/978-3-030-43651-3_58}
Farrell, P., Peschka, D.: Challenges in drift-diffusion semiconductor simulations. In: Kl{\"o}fkorn, R., Keilegavlen, E., Radu, F.A., Fuhrmann, J. (eds.) Finite Volumes for Complex Applications IX - Methods, Theoretical Aspects, Examples. pp. 615--623. Springer International Publishing, Cham (2020)

\bibitem{fernandes2022learning}
Fernandes, P., Treviso, M., Pruthi, D., Martins, A., Neubig, G.: Learning to scaffold: Optimizing model explanations for teaching. Advances in Neural Information Processing Systems  \textbf{35},  36108--36122 (2022)

\bibitem{frazier2016bayesian}
Frazier, P.I., Wang, J.: Bayesian optimization for materials design. Information science for materials discovery and design pp. 45--75 (2016)

\bibitem{garnett_bayesoptbook_2023}
Garnett, R.: {Bayesian Optimization}. Cambridge University Press (2023)

\bibitem{guidotti2018survey}
Guidotti, R., Monreale, A., Ruggieri, S., Turini, F., Giannotti, F., Pedreschi, D.: A survey of methods for explaining black box models. ACM computing surveys (CSUR)  \textbf{51}(5),  1--42 (2018)

\bibitem{7780459}
He, K., Zhang, X., Ren, S., Sun, J.: Deep residual learning for image recognition. In: 2016 IEEE Conference on Computer Vision and Pattern Recognition (CVPR). pp. 770--778 (2016). \doi{10.1109/CVPR.2016.90}

\bibitem{hedstrom2023quantus}
Hedstr{\"o}m, A., Weber, L., Krakowczyk, D., Bareeva, D., Motzkus, F., Samek, W., Lapuschkin, S., H{\"o}hne, M.M.C.: Quantus: An explainable ai toolkit for responsible evaluation of neural network explanations and beyond. Journal of Machine Learning Research  \textbf{24}(34),  1--11 (2023)

\bibitem{Himmelblau1972AppliedNP}
Himmelblau, D.: Applied Nonlinear Programming. McGraw-Hill (1972)

\bibitem{10.5555/3044805.3044891}
Hutter, F., Hoos, H., Leyton-Brown, K.: An efficient approach for assessing hyperparameter importance. In: Proceedings of the 31st International Conference on International Conference on Machine Learning - Volume 32. p. I–754–I–762. ICML'14, JMLR.org (2014)

\bibitem{10.1007/978-3-030-01851-1_23}
Kampman, I., Elomaa, T.: Hierarchical clustering of high-dimensional data without global dimensionality reduction. In: Ceci, M., Japkowicz, N., Liu, J., Papadopoulos, G.A., Ra{\'{s}}, Z.W. (eds.) Foundations of Intelligent Systems. pp. 236--246. Springer International Publishing, Cham (2018)

\bibitem{KLUTCHNIKOFF2022105075}
Klutchnikoff, N., Poterie, A., Rouvière, L.: Statistical analysis of a hierarchical clustering algorithm with outliers. Journal of Multivariate Analysis  \textbf{192},  105075 (2022)

\bibitem{krizhevsky2009learning}
Krizhevsky, A., Hinton, G., et~al.: Learning multiple layers of features from tiny images. Master's thesis, University of Toronto (2009)

\bibitem{liu2022airfoil}
Liu, R.L., Zhao, Q., He, X.J., Yuan, X.Y., Wu, W.T., Wu, M.Y.: Airfoil optimization based on multi-objective bayesian. Journal of Mechanical Science and Technology  \textbf{36}(11),  5561--5573 (2022)

\bibitem{lundberg2019explainable}
Lundberg, S.M., Erion, G.G., Chen, H., DeGrave, A.J., Prutkin, J.M., Nair, B., Katz, R., Himmelfarb, J., Bansal, N., Lee, S.: From local explanations to global understanding with explainable {AI} for trees. Nat. Mach. Intell.  \textbf{2}(1),  56--67 (2020)

\bibitem{macha2022rulexai}
Macha, D., Kozielski, M., Wr{\'{o}}bel, L., Sikora, M.: Rulexai - {A} package for rule-based explanations of machine learning model. SoftwareX  \textbf{20},  101209 (2022)

\bibitem{Magdalena2015}
Magdalena, L.: Fuzzy Rule-Based Systems, pp. 203--218. Springer Berlin Heidelberg, Berlin, Heidelberg (2015)

\bibitem{10066202}
Mahmud, M.S., Huang, J.Z., Ruby, R., Ngueilbaye, A., Wu, K.: Approximate clustering ensemble method for big data. IEEE Transactions on Big Data  \textbf{9}(4),  1142--1155 (2023)

\bibitem{mishra2006some}
Mishra, S.K.: Some new test functions for global optimization and performance of repulsive particle swarm method. Available at SSRN 926132  (2006)

\bibitem{moosbauer2024}
Moosbauer, J., Herbinger, J., Casalicchio, G., Lindauer, M., Bischl, B.: Explaining hyperparameter optimization via partial dependence plots. In: Proceedings of the 35th International Conference on Neural Information Processing Systems. NIPS '21, Curran Associates Inc., Red Hook, NY, USA (2021)

\bibitem{murdoch2017automatic}
Murdoch, W.J., Szlam, A.: Automatic rule extraction from long short term memory networks. arXiv preprint arXiv:1702.02540  (2017)

\bibitem{murtagh2012algorithms}
Murtagh, F., Contreras, P.: Algorithms for hierarchical clustering: an overview. Wiley Interdisciplinary Reviews: Data Mining and Knowledge Discovery  \textbf{2}(1),  86--97 (2012)

\bibitem{murtagh2011ward}
Murtagh, F., Legendre, P.: Ward's hierarchical clustering method: Clustering criterion and agglomerative algorithm. CoRR  \textbf{abs/1111.6285} (2011)

\bibitem{nagataki2022online}
Nagataki, M., Kondo, K., Yamazaki, O., Yuki, K., Nakazawa, Y.: Online auto-tuning method in field-orientation-controlled induction motor driving inertial load. IEEE Open Journal of Industry Applications  \textbf{3},  125--140 (2022)

\bibitem{10.1145/3583558}
Nauta, M., Trienes, J., Pathak, S., Nguyen, E., Peters, M., Schmitt, Y., Schl\"{o}tterer, J., van Keulen, M., Seifert, C.: From anecdotal evidence to quantitative evaluation methods: A systematic review on evaluating explainable ai. ACM Computing Surveys  (2023)

\bibitem{neumann2019data}
Neumann-Brosig, M., Marco, A., Schwarzmann, D., Trimpe, S.: Data-efficient autotuning with bayesian optimization: An industrial control study. IEEE Transactions on Control Systems Technology  \textbf{28}(3),  730--740 (2019)

\bibitem{ortiz2024enhanced}
Ortiz, R., Miranda-Chiquito, P., Encalada-Davila, A., Marquez, L.E., Tutiven, C., Chatzi, E., Silva, C.E.: An enhanced modeling framework for bearing fault simulation and machine learning-based identification with bayesian-optimized hyperparameter tuning. Journal of Computing and Information Science in Engineering  \textbf{24}(9),  091002 (2024)

\bibitem{pahde2023optimizing}
Pahde, F., Yolcu, G.{\"U}., Binder, A., Samek, W., Lapuschkin, S.: Optimizing explanations by network canonization and hyperparameter search. In: Proceedings of the IEEE/CVF Conference on Computer Vision and Pattern Recognition. pp. 3818--3827 (2023)

\bibitem{ran2023comprehensive}
Ran, X., Xi, Y., Lu, Y., Wang, X., Lu, Z.: Comprehensive survey on hierarchical clustering algorithms and the recent developments. Artificial Intelligence Review  \textbf{56}(8),  8219--8264 (2023)

\bibitem{Rasmussen2005-yj}
Rasmussen, C.E., Williams, C.K.I.: Gaussian processes for machine learning. Adaptive computation and machine learning, {MIT} Press (2006)

\bibitem{ribeiro2016why}
Ribeiro, M.T., Singh, S., Guestrin, C.: "why should {I} trust you?": Explaining the predictions of any classifier. In: Proceedings of the International Conference on Knowledge Discovery and Data Mining, {SIGKDD}. pp. 1135--1144. {ACM} (2016)

\bibitem{rodemann2022accounting}
Rodemann, J., Augustin, T.: Accounting for gaussian process imprecision in bayesian optimization. In: International Symposium on Integrated Uncertainty in Knowledge Modelling and Decision Making. pp. 92--104. Springer (2022)

\bibitem{rodemann2024explaining}
Rodemann, J., Croppi, F., Arens, P., Sale, Y., Herbinger, J., Bischl, B., Hüllermeier, E., Augustin, T., Walsh, C.J., Casalicchio, G.: Explaining bayesian optimization by shapley values facilitates human-ai collaboration (2024)

\bibitem{roussel2024bayesian}
Roussel, R., Edelen, A.L., Boltz, T., Kennedy, D., Zhang, Z., Ji, F., Huang, X., Ratner, D., Garcia, A.S., Xu, C., et~al.: Bayesian optimization algorithms for accelerator physics. Physical Review Accelerators and Beams  \textbf{27}(8),  084801 (2024)

\bibitem{rudin2022black}
Rudin, C.: Why black box machine learning should be avoided for high-stakes decisions, in brief. Nature Reviews Methods Primers  \textbf{2}(1), ~81 (2022)

\bibitem{salvador2004determining}
Salvador, S., Chan, P.: Determining the number of clusters/segments in hierarchical clustering/segmentation algorithms. In: 16th IEEE international conference on tools with artificial intelligence. pp. 576--584. IEEE (2004)

\bibitem{schwalbe2023comprehensive}
Schwalbe, G., Finzel, B.: A comprehensive taxonomy for explainable artificial intelligence: a systematic survey of surveys on methods and concepts. Data Mining and Knowledge Discovery pp. 1--59 (2023)

\bibitem{segel23a}
Segel, S., Graf, H., Tornede, A., Bischl, B., Lindauer, M.: Symbolic explanations for hyperparameter optimization. In: Faust, A., Garnett, R., White, C., Hutter, F., Gardner, J.R. (eds.) Proceedings of the Second International Conference on Automated Machine Learning. Proceedings of Machine Learning Research, vol.~224, pp. 2/1--22. PMLR (12--15 Nov 2023)

\bibitem{seitz2022gradient}
Seitz, S.: Gradient-based explanations for gaussian process regression and classification models. arXiv preprint arXiv:2205.12797  (2022)

\bibitem{shahriari2015taking}
Shahriari, B., Swersky, K., Wang, Z., Adams, R.P., De~Freitas, N.: Taking the human out of the loop: A review of bayesian optimization. Proceedings of the IEEE  \textbf{104}(1),  148--175 (2015)

\bibitem{silagadze2007finding}
Silagadze, Z.: Finding two-dimensional peaks. physics of Particles and Nuclei Letters  \textbf{4},  73--80 (2007)

\bibitem{snoek2012practical}
Snoek, J., Larochelle, H., Adams, R.P.: Practical bayesian optimization of machine learning algorithms. Advances in neural information processing systems  \textbf{25} (2012)

\bibitem{sundin2022human}
Sundin, I., Voronov, A., Xiao, H., Papadopoulos, K., Bjerrum, E.J., Heinonen, M., Patronov, A., Kaski, S., Engkvist, O.: Human-in-the-loop assisted de novo molecular design. Journal of Cheminformatics  \textbf{14}(1),  1--16 (2022)

\bibitem{VELLIANGIRI2019104}
Velliangiri, S., Alagumuthukrishnan, S., {Thankumar joseph}, S.I.: A review of dimensionality reduction techniques for efficient computation. Procedia Computer Science  \textbf{165},  104--111 (2019)

\bibitem{vilone2021quantitative}
Vilone, G., Longo, L.: A quantitative evaluation of global, rule-based explanations of post-hoc, model agnostic methods. Frontiers in artificial intelligence  \textbf{4},  717899 (2021)

\bibitem{van2021evaluating}
van~der Waa, J., Nieuwburg, E., Cremers, A.H.M., Neerincx, M.A.: Evaluating {XAI:} {A} comparison of rule-based and example-based explanations. Artif. Intell.  \textbf{291},  103404 (2021)

\bibitem{wang2015falling}
Wang, F., Rudin, C.: Falling rule lists. In: Artificial intelligence and statistics. pp. 1013--1022. PMLR (2015)

\bibitem{ward1963hierarchical}
Ward~Jr, J.H.: Hierarchical grouping to optimize an objective function. Journal of the American statistical association  \textbf{58}(301),  236--244 (1963)

\bibitem{10.1007/978-3-662-64408-9_7}
Wirth, C., Schmid, U., Voget, S.: Humanzentrierte k{\"u}nstliche intelligenz: Erkl{\"a}rendes interaktives maschinelles lernen f{\"u}r effizienzsteigerung von parametrieraufgaben. In: Hartmann, E.A. (ed.) Digitalisierung souver{\"a}n gestalten II. pp. 80--92. Springer Berlin Heidelberg, Berlin, Heidelberg (2022)

\bibitem{WITTEN201767}
Witten, I.H., Frank, E., Hall, M.A., Pal, C.J.: Chapter 3 - output: Knowledge representation. In: Data Mining (Fourth Edition), pp. 67--89. Morgan Kaufmann, fourth edition edn. (2017)

\bibitem{wu2020prediction}
Wu, Z., et~al.: Prediction of california house price based on multiple linear regression. Academic Journal of Engineering and Technology Science  \textbf{3}(7.0) (2020)

\bibitem{xu2008uncertainty}
Xu, C., Gertner, G.Z.: Uncertainty and sensitivity analysis for models with correlated parameters. Reliability Engineering \& System Safety  \textbf{93}(10),  1563--1573 (2008)

\bibitem{yazgana2016literature}
Yazgana, P., Kusakci, A.O.: A literature survey on association rule mining algorithms. Southeast Europe Journal of soft computing  \textbf{5}(1) (2016)

\bibitem{zhou2021evaluating}
Zhou, J., Gandomi, A.H., Chen, F., Holzinger, A.: Evaluating the quality of machine learning explanations: A survey on methods and metrics. Electronics  \textbf{10}(5), ~593 (2021)

\end{thebibliography}





\end{document}